\documentclass[11pt, oneside]{article}   	
\usepackage[margin=1.0in]{geometry}                		
\usepackage{algorithm}
\usepackage{algpseudocode}
\usepackage{amsmath}
\usepackage{authblk}
\usepackage{bm}
\usepackage{float}
\usepackage{graphicx}				
\usepackage{hyperref}	
\hypersetup{linkcolor=blue, colorlinks=true, urlcolor=blue, filecolor=blue, citecolor=blue}		
\usepackage{listings}								
\usepackage{comment}
\usepackage{caption}
\usepackage{subcaption}
	
\usepackage{xcolor}
\usepackage{soul}
\usepackage{amssymb}
\usepackage[numbers,square]{natbib}
\title{An out-of-distribution discriminator based on Bayesian neural network epistemic uncertainty}
\author[1]{\href{mailto:ancell@uw.edu}{Ethan Ancell}}
\author[2]{\href{mailto:cbennet@sandia.gov}{Christopher Bennett}}
\author[2]{\href{mailto:bjdebus@sandia.gov}{Bert Debusschere}}
\author[2]{\href{mailto:sagarwa@sandia.gov}{Sapan Agarwal}}
\author[2]{\href{mailto:phays@sandia.gov}{Park Hays}}
\author[2]{\href{mailto:txiao@sandia.gov}{T. Patrick Xiao}}
\affil[1]{Department of Statistics, University of Washington}
\affil[2]{Sandia National Laboratories}
\date{}	
\begin{document}						
\maketitle

\section{Abstract}

Neural networks have revolutionized the field of machine learning with increased predictive capability over traditional methods. In addition to the demand for improved predictions, there is a simultaneous demand for reliable uncertainty quantification on estimates made by machine learning methods such as neural networks. Bayesian neural networks (BNNs) are an important type of neural network with built-in capability for quantifying uncertainty. This paper begins with a review of aleatoric and epistemic uncertainty in BNNs, then relates these concepts with an image generator which creates a dataset where the goal is to identify the amplitude of an event in the image. It is shown that epistemic uncertainty tends to be lower in images which are well-represented in the training dataset and tends to be high in images which are not well-represented. An algorithm for out-of-distribution (OoD) detection with BNN epistemic uncertainty is introduced along with various experiments demonstrating factors influencing the OoD detection capability in a BNN. The OoD detection capability with epistemic uncertainty is compared with the OoD detection in the discriminator network of a generative adversarial network (GAN) with comparable network architecture.

\section{Introduction}
\label{sec:intro}

Uncertainty quantification (UQ) is important for ensuring trust in the predictions made by machine learning algorithms but is seldom provided as a feature in conventional deep neural networks (DNNs). Without reliable UQ, DNN predictions cannot be relied upon for high-consequence decisions, because their trustworthiness cannot be evaluated in a context where the cost of an incorrect prediction is high. These safety-critical contexts include the use of machine learning for medical diagnoses \citep{begoli2019need}, self-driving cars, physical infrastructure (e.g. smart grid) \citep{varshney2017safety}, national security, and other applications. An important component for safety is out-of-distribution (OoD) detection, where the neural network identifies inputs as being sufficiently unlikely to be part of the distribution of training data, implying that any prediction made on that input should not be trusted. When studying OoD detection using UQ methods in machine learning, we will reference two different types of uncertainties:
\begin{enumerate}
	\item \textit{Aleatoric uncertainty}: Uncertainty inherent in the data that cannot be reduced with more data.
	\item \textit{Epistemic uncertainty}: Uncertainty in the model parameters that is reduced with more training examples.c
\end{enumerate}
Practically, OoD data should consist of inputs for which the neural network has high epistemic uncertainty, because epistemic uncertainty could have been reduced by including the OoD data inside the training set \citep{d2021out}. 

Bayesian neural networks (BNNs) are a variant of DNNs where the parameters are random variables rather than fixed values. Training BNNs involves using Bayes’ theorem to infer probability distributions over their parameters based on observed training data and prior knowledge. BNNs can provide better calibrated uncertainty estimates than conventional DNNs, and by representing weights as random variables, BNNs are capable of quantifying epistemic uncertainty. Aleatoric uncertainty can also be estimated with BNNs: this uncertainty is typically obtained by training a BNN to explicitly predict the aleatoric uncertainty rather than relying on the stochasticity of the model parameters themselves \citep{kendall2017uncertainties}. The ability of BNNs to predict both types of uncertainties has made BNNs attractive for UQ, and more specifically they have been studied for the problem of OoD detection. 

Some previous literature in OoD detection with BNNs has studied this problem using the framework of Gaussian processes (GPs), because infinite-width BNNs with Gaussian priors over the weights are equivalent to GPs for this problem \citep{neal1996priors}. The paper \citep{guenais2020bacoun} tackles OoD detection using Neural Linear Models (NLMs) and an augmentation of the data with additional points lying on the periphery of the training data and connects this with OoD detection with GPs and BNNs. The reference \citep{mitros2020ramifications} shows how BNNs can be used for OoD detection using the max probability, entropy, mutual information, and differential entropy from the output vector of the DNN and shows that the BNNs outperform DNNs trained for the same task. Others have criticized the use of BNNs for OoD detection, arguing that OoD detection with BNNs is sensitive to the choice of prior over the weights and may have a trade-off with generalization \citep{henning2021bayesian}. In this paper, we expand upon these results by examining the problem of OoD detection with BNNs under a simplistic framework where inputs are marked as in-distribution or out-of-distribution based upon previously seen values of epistemic uncertainty in a validation dataset. Instead of using max probability, entropy, mutual information, or differential entropy for OoD detection as in \citep{mitros2020ramifications}, we base our discriminator on epistemic uncertainty as introduced in \citep{gal2016} for a simple regression setting.

We train BNNs for regression on a dataset of event images with a well-defined, parameterized generating function. The BNNs are tested on OoD test inputs that are formed by corrupting in-distribution (iD) images with noise that is uncharacteristic of the training set, or by superimposing iD images with images from a different dataset. We compare the OoD detection capability of BNNs to that of Generative Adversarial Networks (GANs). The discriminator network in a GAN is rewarded for detecting the outputs of a generator network as OoD, while the generator network is rewarded for fooling the discriminator. The GAN is therefore naturally suited to the task of OoD detection and is a useful comparison point for BNN-based OoD detection, despite the difference in the detection mechanism.

This paper begins with a brief review of Bayesian statistical methods and BNNs in Section \ref{sec:bayesian_review}. Section \ref{sec:data_description} introduces the dataset of event images with well-defined properties that are used to train the networks in this study. Section \ref{sec:calculating_uncertainty} relates aleatoric and epistemic uncertainty to qualitative differences in various images from the motivating dataset. Section \ref{sec:comparison_with_GAN} demonstrates that the epistemic uncertainty predicted by BNNs can be used to detect OoD inputs and compares the approach to OoD detection using the discriminator network in a GAN. Overall, we find that the BNN epistemic uncertainty approach to OoD detection has similar sensitivity to the GAN approach. The advantage of the BNN is that the OoD detection capability is built into the same model that is trained to accurately perform the original neural network task (e.g. regression).

The contributions of this paper are as follows:
\begin{itemize}
	\item We quantify the difference in the predicted epistemic uncertainty of a BNN on in-distribution vs out-of-distribution inputs by using a dataset with a well-defined generating function and different methods of generating OoD data.
	\item We propose a simple and robust algorithm based on BNN epistemic uncertainty to classify inputs as in-distribution or out-of-distribution.
	\item We investigate how the effectiveness of OoD detection depends on the size of the training set, epochs used for training, and the topology of the BNN.
	\item We show that OoD detection using epistemic uncertainty achieves similar sensitivity as a GAN discriminator of similar complexity.
\end{itemize}

\section{Review of Bayesian methods and neural networks}
\label{sec:bayesian_review}

Traditional neural networks carry fixed weights and biases. In Bayesian neural networks (BNNs), the weights and biases are random variables. A forward pass is conducted by randomly sampling the weights and biases from their distributions and using the sampled values in the forward pass. The distributions of weights and biases in a Bayesian neural network are updated using algorithms from Bayesian statistics.

In Bayesian methods, a prior $p(\theta)$ is posited that represents prior knowledge about the collection of parameters $\theta$ of the network. After collecting data $\mathcal{D}$, the prior distribution is combined with the likelihood of the data (which accounts for its associated uncertainty) $p(\mathcal{D} ~ \vert ~ \theta)$ to obtain a posterior distribution $p(\theta ~ \vert ~ \mathcal{D})$ using Bayes Theorem:
\begin{equation}
	p(\theta ~ \vert ~ \mathcal{D}) = \dfrac{p(\mathcal{D} ~ \vert ~ \theta)p(\theta)}{p(\mathcal{D})}
	\label{eq:bayes}
\end{equation}
This paper will work under the supervised learning paradigm: $\mathcal{D}$ refers to a training dataset as a complete entity where $\mathcal{D} = (\bm{X}, \bm{y})$. The matrix $\bm{X} \in \mathcal{X} \subset \mathbb{R}^{n \times d}$ is the training data with size $n$, dimensionality $d$, and support $\mathcal{X}$. The vector $\bm{y} \in \mathcal{Y} \subset \mathbb{R}^{n}$ is the set of real-valued labels with support $\mathcal{Y}$ corresponding to the training data $\bm{X}$. The goal in a supervised regression task is to find $p(y ~ \vert ~ x)$, and use this conditional distribution to create an optimal prediction $\hat{y}$ (e.g., with the posterior mean). The neural network will be written as a function $\Phi: \mathcal{X} \to \mathcal{Y}$. Writing $\Phi$ to have explicit dependence on its parameters, a forward pass through the BNN is written $\Phi(x ~ \vert ~ \theta_i)$ where $\theta_i$ is a realization of the parameters from the posterior distribution $p(\theta ~ \vert ~ \mathcal{D})$.

Updating the weights and biases in a network is nontrivial: certain components of Equation \ref{eq:bayes} are mathematically intractable to calculate in a practical setting, especially the evidence $p(\mathcal{D})$ which would require an expensive numerical integration of the numerator of (\ref{eq:bayes}) over the high-dimensional parameter space of $\theta$. Given the intractability of using Bayes Theorem directly, there are two popular workaround methods for sampling from the posterior $p(\theta ~ \vert ~ \mathcal{D})$: Markov Chain Monte Carlo (MCMC) and Variational Inference (VI). MCMC algorithms can produce a near-exact sampling from the posterior $p(\theta ~ \vert ~ \mathcal{D})$, but they are not computationally scalable and hence are impractical for large-scale BNNs \citep{jospin2022}.

Variational inference algorithms posit a class of distributions $Q$ that are used to approximate the posterior $p(\theta ~ \vert ~ \mathcal{D})$. The goal of a variational inference algorithm is to find an optimal distribution $q^*(\theta) \in Q$ such that $q^*(\theta)$ is most similar to the true posterior distribution $p(\theta ~ \vert ~ \mathcal{D})$. Similarity in this context is measured with the Kullback-Leibler (KL) divergence between a given $q(\theta) \in Q$ and the posterior $p(\theta ~ \vert ~ \mathcal{D})$. Minimizing the KL divergence directly is difficult (it would require the computation of the evidence), but the KL divergence can be manipulated to separate out and neglect the evidence from the optimization problem, since it is independent of $q(\theta)$. The solution to the variational inference problem using this modified objective function, called the evidence lower bound (ELBO), is equivalent to solving the original problem up to an additive constant. 

The BNNs in this paper are trained with the Flipout method for efficient mini-batch optimization \citep{wen2018flipout}. Unless otherwise stated, all neural network models mentioned in this paper will be BNNs with the prior distributions for the weights being independent standard normal distributions: so all means and standard deviations describing the weights are instantiated with mean parameter 0 and standard deviation 1. The approximating class of distributions used in variational inference will be the class of independent normal distributions where the means and standard deviations of these distributions are inferred with training. We use TensorFlow and the TensorFlow Probability Python libraries to create and test these models.

\noindent
Methods for calculating aleatoric and epistemic uncertainty in a BNN will be described in Section \ref{sec:calculating_uncertainty} after an introduction to the motivating dataset.

\section{The amplitudes dataset}
\label{sec:data_description}

This paper will repeatedly reference an ``amplitudes dataset,'' a dataset consisting of $40 \times 40$ grayscale images which may contain one or no events. An ``event'' is generated by a point spread function $\text{PSF}(A,x,y)$ superimposed on a noisy background, with each event having a specified amplitude $A$ and center coordinate $(x,y)$ within the image. Without an event, the image consists of noise only. This dataset emulates a simple, hypothetical anomaly detection application based on image sensor data. The generated images can be passed into a neural network that predicts the presence, amplitude, and coordinates of an event, using the true values as labels. In this paper, the networks will only be trained to predict the amplitude of an image. Nonevent images are handled by treating them as having an amplitude of 0.

\begin{figure}[H]
	\centering
	\includegraphics[width=1.0\textwidth]{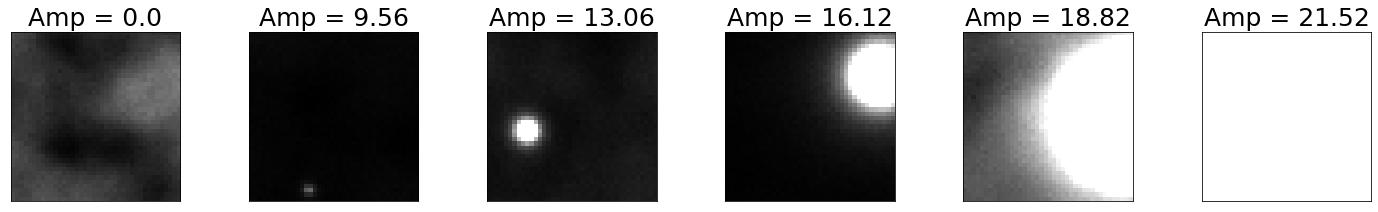}
	\caption{A few different amplitude levels in the dataset}
	\label{fig:varying_amp_levels}
\end{figure}

\begin{figure}[H]
	\centering
	\includegraphics[width=0.50\textwidth]{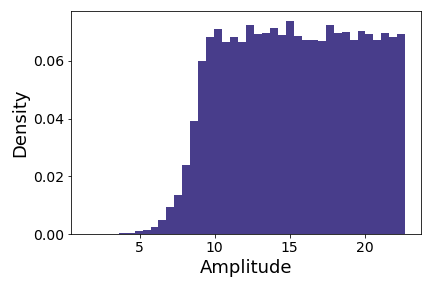}
	\caption{Histogram for the distribution of amplitudes in the dataset. This histogram represents only the event images, which is 50\% of the data. The other 50\% are nonevent images and thus have zero amplitude.}
	\label{fig:amplitudes_distribution}
\end{figure}

For a high amplitude event, the PSF generates a signal that has higher pixel brightness values and a larger spatial extent within the image. The PSF also models sensor saturation by limiting brightness values to the range $[0,1]$. A bright event can cause many pixels to saturate over a large area. Figure \ref{fig:varying_amp_levels} shows example images from the dataset.
Figure \ref{fig:amplitudes_distribution} shows the distribution of amplitudes among all images in the training set. In this dataset, images with an amplitude over approximately 21 will fully saturate the image, as is in the rightmost image shown in Figure \ref{fig:varying_amp_levels}.
\noindent
The dataset is synthetically generated, so the user can manually control parameters such as the proportion of images with an event, the size of the training and test sets, and the resolution of the image. Unless otherwise stated, all experiments will be trained with a training size of 10000 images with $40 \times 40$ resolution where there is a 50\% split in event/non-event images. In this section, the symbols $A, x, y$ referred to amplitude, $x$-position, and $y$-position respectively. The remainder of this paper will only concern the prediction of amplitude, so the symbol $y$ will supplant $A$ for the amplitude of an image as to conform with standard labeling notation in machine learning.

The value of the amplitudes dataset lies in the control that the user has in adjusting the generation parameters, making it ideal for studying how specific perturbations to the training and testing datasets creates downstream differences in the characteristics of UQ using BNNs. Furthermore, the dataset is sufficiently complex such that the insights in this paper are likely generalizable to other image dataset problems.

\section{Calculating uncertainty}
\label{sec:calculating_uncertainty}

Recall from Section \ref{sec:intro} that aleatoric uncertainty is the uncertainty inherent in the data, and epistemic uncertainty is the uncertainty in the model parameters. There are varying approaches to calculate a number for these uncertainties for a trained model. The sections below describe how to calculate aleatoric and epistemic uncertainty in a BNN. As notation, $\hat{\sigma}_A$ and $\hat{\sigma}_E$ will represent the estimated aleatoric and epistemic uncertainty respectively.

\subsection{Calculating aleatoric uncertainty}
\label{sec:calc_alea}

Aleatoric uncertainty in BNNs is estimated by including it as an explicit prediction of the neural network. Each forward pass through the network results in a tuple $(\hat{y}, \hat{\sigma}_A)$ where $\hat{\sigma}_A$ is a numeric representation of the aleatoric uncertainty in the prediction $\hat{y}$. This method of calculating aleatoric uncertainty was originally proposed by \cite{kendall2017uncertainties}. Allowing the network to make varying predictions of $\hat{\sigma}_A$ for different inputs allows the model to capture heteroscedastic (e.g., non-constant and dependent on the input) noise in the data. For a full discussion on why a heteroscedastic noise model for aleatoric uncertainty is appropriate in the amplitudes dataset, see Appendix \ref{app:heteroscedastic_evidence}.

There is only one label $y \in \mathbb{R}$ for the regression task of predicting the amplitude of an image, but the prediction of $\hat{\sigma}_A$ can be included in the loss function of the network if the loss function is the negative log-likelihood of a Gaussian distribution. This choice of loss function implicitly assumes an approximate Gaussian spread of $Y$ around $\Phi(X ~|~ \theta_t)$ with a scale parameter of $\sigma_A^2$ (note we could write $\sigma_A^2 = \sigma_A^2(X)$ to make the heteroscedastic noise model on $Y$ more explicit). Letting $\hat{y}$ be the location parameter and $\hat{\sigma}_A$ be the scale parameter, the negative log-likelihood loss function for a single prediction $(\hat{y}, \hat{\sigma}_A)$ with true label $y$ is given by
\begin{equation}
	-\ell (\hat{y}, \hat{\sigma}_A ~ \vert ~ y) = -\log \left( \hat{\sigma}_A^{-1} (2 \pi)^{-1/2} \exp \left( \dfrac{-(\hat{y} - y)^2}{2 \hat{\sigma}_A^2} \right) \right)
\end{equation}
Writing $\hat{\bm{y}} = (\hat{y}_1, \hat{y}_2, \dots, \hat{y}_n)$ as the set of all predictions for a data matrix $\bm{X}$ with true labels $\bm{y}$, the aggregate loss function is the joint negative log-likelihood:
\begin{equation}
	\text{loss}(\hat{\bm{y}}, \bm{y}) = -\sum_{i=1}^{n} \ell(\hat{y}_i, \hat{\sigma}_{A,i} ~ \vert ~ y_i) = -\sum_{i=1}^{n} \log \left( \hat{\sigma}_{A,i}^{-1} (2 \pi)^{-1/2} \exp \left( \dfrac{(\hat{y}_i - y_i)^2}{2\hat{\sigma}_{A,i}^2} \right) \right)
	\label{eq:nll_loss}
\end{equation}
Equation (\ref{eq:nll_loss}) can be further simplified to:
\begin{align*}
	\text{loss}(\hat{\bm{y}}, \bm{y}) = -\sum_{i=1}^{n} \dfrac{(\hat{y}_i - y_i)^2}{2 \hat{\sigma}_{A,i}^2} + (2\pi)^{1/2} \sum_{i=1}^{n} \log \hat{\sigma}_{A,i}
\end{align*}
The purpose of this simplification is to emphasis that the joint negative log-likelihood loss function contains a natural regularization on $\hat{\sigma}_A$ in the second term of the above expression, addressing a concern that $\hat{\sigma}_A$ might explode during training so that poor predictions still have high likelihood.

A distribution-inspired loss function implicitly accommodates uncertainty in the placement of $y$ around $\hat{y}$. The parameterization of this uncertainty with $\hat{\sigma}_A$ from the network results in a useful approach for estimating aleatoric uncertainty. Note that some recent papers have discovered pitfalls with the negative log-likelihood loss function above, such as the reference \cite{seitzer2022pitfalls} which claims that in certain specific data situations, the natural regularization on $\hat{\sigma}_A$ in Equation \ref{eq:nll_loss} may not be sufficient. We accept a possible small deviation from optimality in our BNNs stemming from this fact, and do not believe that the amplitudes dataset falls into the edge cases mentioned in this paper.

As a technical note on implementation, given that $\hat{\sigma}_A$ should always be positive for statistical reasons it is best to use a softplus transformation $\text{sp}: \mathbb{R} \to \mathbb{R}^+$ whenever predicting $\hat{\sigma}_A$ for use in the loss function or generating confidence intervals.
\[\text{sp}(t) = \log(1 + \exp(t))\]

\subsection{Calculating epistemic uncertainty}
\label{sec:calc_epis}

Epistemic uncertainty captures the uncertainty in the model parameters. Uncertainty in the trained model parameters is captured by the posterior $p(\theta ~ \vert ~ \mathcal{D})$. The impact of epistemic uncertainty on predictions is obtained by resampling the network, an approach introduced by \citep{kendall2017uncertainties}. This approach is also known as computing the posterior push-forward distribution.

A set of sampled parameters $\theta_1, \theta_2, \dots, \theta_T \sim p(\theta ~ \vert ~ \mathcal{D})$ are independent and identically distributed random variables, so the network predictions conditional on a fixed $X$ will also be independent and identically distributed random variables for different realizations of $\theta$. The calculation for epistemic uncertainty $\hat{\sigma}_E$ is given by the sample standard deviation of the posterior push-forward distribution, or equivalently the sample standard deviation of the realizations of the random variables $\Phi(x ~ \vert ~ \theta_1), \Phi(x ~ \vert ~ \theta_2), \dots, \Phi(x ~ \vert ~ \theta_T)$:

\[\text{resamples} = \{\hat{y}_{i}\}_{i=1}^T = \{\Phi(x ~ \vert ~ \theta_i)\}_{i=1}^{T}, \quad \quad \bar{y} = \dfrac{1}{T} \sum_{i=1}^T \hat{y}_i\]
\begin{equation}
	\hat{\sigma}_E = \sqrt{ \dfrac{1}{T-1} \sum_{i=1}^T (\hat{y}_i - \bar{y})^2 }
	\label{eq:calc_epis}
\end{equation}
When $\hat{\sigma}_E$ is small, the distribution of $\Phi(x ~ \vert ~ \theta_i)$ has less spread, so there is less uncertainty in $p(\theta ~ \vert ~ \mathcal{D})$. Conversely, when $\hat{\sigma}_E$ is high, this indicates that the uncertainty in the predictions is also high.

\subsection{Examples of aleatoric and epistemic uncertainty in the amplitudes dataset}
\label{sec:ex_alea_epis}

\begin{figure}[H]
	\centering
	\includegraphics[width=0.55\textwidth]{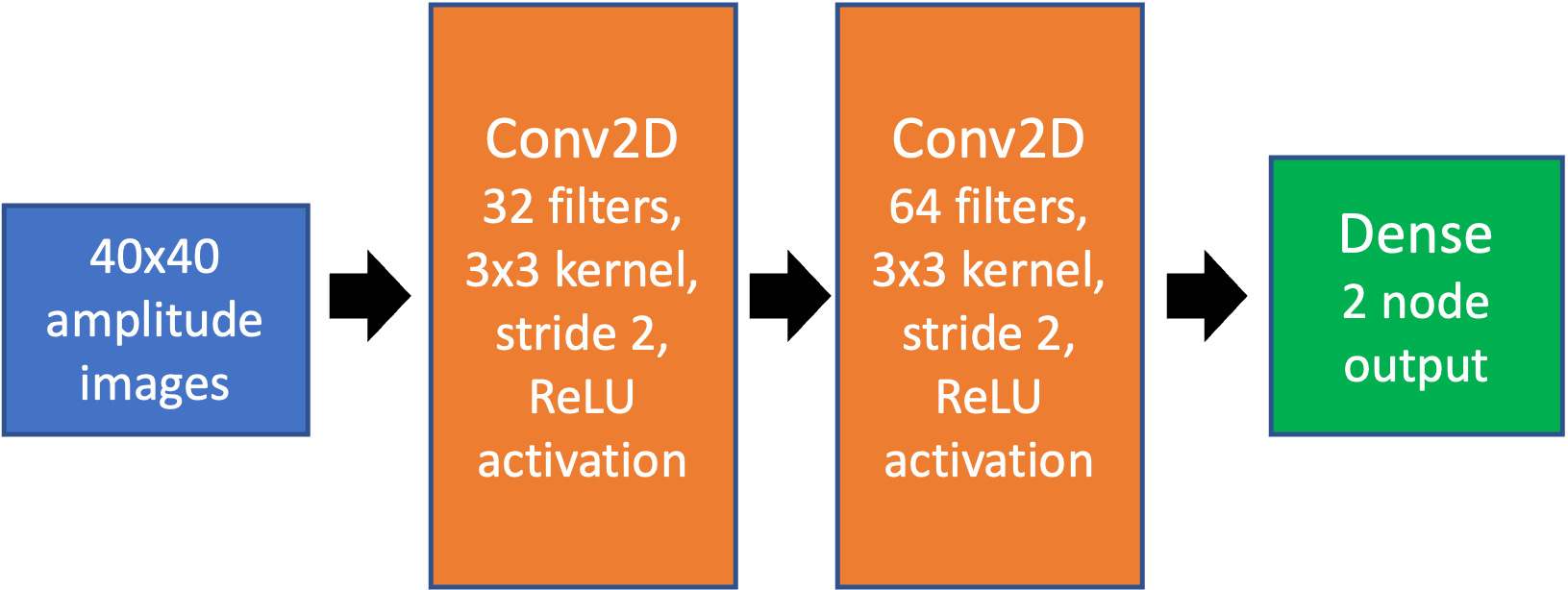}
	\caption{Architecture for Model 1. This model uses a batch size of 64, default TensorFlow probability Adam optimizer with learning rate of 0.0001, 600 training epochs, and 10000 training images.}
	\label{fig:MODEL_1}
\end{figure}

\begin{figure}[H]
	\centering
	\begin{subfigure}[b]{0.40\textwidth}
		\centering
		\includegraphics[width=\textwidth]{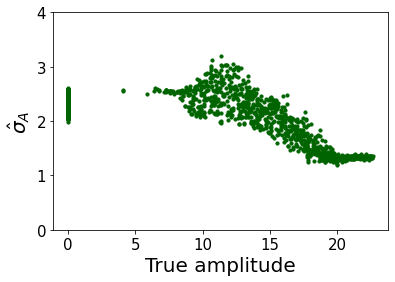}
		\caption{}
		\label{fig:amp_vs_alea}
	\end{subfigure}
	\begin{subfigure}[b]{0.40\textwidth}
		\centering
		\includegraphics[width=\textwidth]{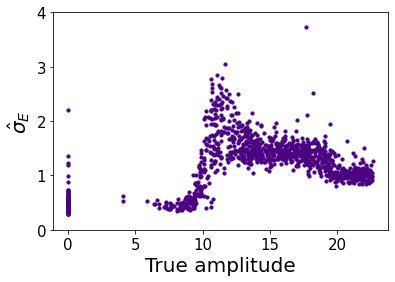}
		\caption{}
		\label{fig:amp_vs_epis}
	\end{subfigure}
	\begin{subfigure}[b]{0.40\textwidth}
		\centering
		\includegraphics[width=\textwidth]{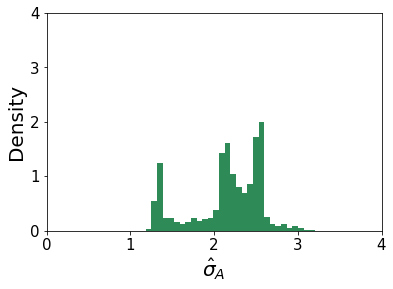}
		\caption{}
		\label{fig:alea_dist}
	\end{subfigure}
	\begin{subfigure}[b]{0.40\textwidth}
		\centering
		\includegraphics[width=\textwidth]{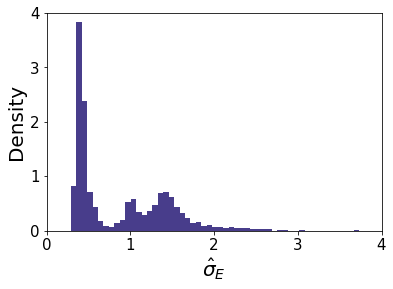}
		\caption{}
		\label{fig:epis_dist}
	\end{subfigure}
	
	\caption{Plots of true amplitude against $\hat{\sigma}_A$ and $\hat{\sigma}_E$, as well as distributions of $\hat{\sigma}_A$ and $\hat{\sigma}_E$.}
	\label{fig:ae_dists_scatter}
\end{figure}

With the amplitudes dataset described in Section \ref{sec:data_description} with a training size of 10000 images and model architecture in Figure \ref{fig:MODEL_1}, Figure \ref{fig:ae_dists_scatter} shows $\hat{\sigma}_A$ and $\hat{\sigma}_E$ as functions of true image amplitude, along with the distributions of predicted $\hat{\sigma}_A$ and $\hat{\sigma}_E$ on a testing dataset of 2000 images. To calculate both $\hat{\sigma}_A$ and $\hat{\sigma}_E$ for a given image, a resampling parameter of $T=100$ in the BNN was used.

On a test dataset of 2000 withheld images, this trained model yielded an RMS amplitude error of 2.0218. The top row of Figure \ref{fig:ae_dists_scatter} shows plots of $\hat{\sigma}_A$ and $\hat{\sigma}_E$ as a function of true image amplitude, and the bottom row of Figure \ref{fig:ae_dists_scatter} shows the distributions of $\hat{\sigma}_A$ and $\hat{\sigma}_E$ on this test dataset.

In Figure \ref{fig:amp_vs_alea}, note that images with high aleatoric uncertainty are primarily non-event and low amplitude images. Images with low amplitudes have low signal-to-noise ratio; in general, it is difficult to tell whether the difference in two images comes from different event amplitudes (e.g. 0 and 10: see Figure \ref{fig:varying_amp_levels}) or differences in the random noise. Therefore, there is significant uncertainty inherent in the data on the true amplitude of the event. Images with lower aleatoric uncertainty lie in a certain range of the data where the amplitude of the event is well-predicted, due to a tendency for higher-amplitude images to have a clear outline of where the end of the event lies inside the imaging array. Because there is not much uncertainty in the amplitude for this type of image, it has a low value of $\hat{\sigma}_A$.

Figure \ref{fig:amp_vs_epis} shows that the epistemic uncertainty varies with amplitude in a manner distinct from aleatoric uncertainty. The epistemic uncertainty depends primarily on the likelihood that a test image chosen at a given amplitude will be similar to an image that occurs in the training set. All images that lack an event are broadly similar due to the similarity of the noise background, and are also well represented in the training set (50\% of data), so these images are likely to have low $\hat{\sigma}_E$. All images with low amplitude ($<$10) also look similar because the event is only faintly visible above the noise background. Images with amplitudes in the range between 10 and 15 have the highest visual diversity within the dataset, because the event is readily distinguishable from noise and occupies a small enough portion of the image that different event coordinates lead to highly distinct images (see Figure \ref{fig:varying_amp_levels}). A test image in this range is most likely to differ significantly from any image seen in the training set, leading to high epistemic uncertainty. Images with high amplitude ($>$16) have reduced epistemic uncertainty because these events tend to saturate a large cluster of pixels. This causes images with events in different coordinates to appear visually similar, reducing diversity.

An additional empirical study on aleatoric and epistemic uncertainty as a function of noise levels in the input images can be found in Appendix \ref{app:iD_noise_levels}. This study varies the level of background noise in the amplitudes dataset generator and reports the trends in aleatoric and epistemic uncertainty. The results show that as background image noise increases, calculated aleatoric uncertainty in a BNN increases but epistemic uncertainty remains relatively constant. The purpose of this study is to provide an extra argument that aleatoric and epistemic uncertainty as discussed in Section \ref{sec:calculating_uncertainty} really do capture the respective notions of uncertainty inherent in the training dataset and uncertainty in the model parameters.

\section{Out-of-distribution (OoD) detection with epistemic uncertainty}
\label{sec:ood_with_epistemic}

Epistemic uncertainty is the uncertainty in the model parameters, and can be reduced by providing more training data. Because epistemic uncertainty is reduced with more training data, epistemic uncertainty for an image should decrease if a larger number of similar types of images exist in the training dataset.

This section explores the use of epistemic uncertainty for detecting out-of-distribution (OoD) (i.e., highly unlikely) images. The advantage of OoD detection using epistemic uncertainty is that it does not require the training of a separate network, nor does it require any prior knowledge of the probability density function for the training data. The paper \citep{sedlmeier2019uncertainty} shows that epistemic uncertainty can be used to differentiate between in-distribution and out-of-distribution samples in reinforcement learning applications. Other results have shown that in Bayesian methods, the epistemic uncertainty for an input $x$ is approximately inversely proportional to the density function $p(x)$ \citep{oaza1996bayesian}. This reference however does not explore this claim for BNNs specifically; see Appendix \ref{app:pdf_inverse} for an experiment with evidence for this claim in the case of a BNN.

If the probability density function for an input is inversely proportional to the epistemic uncertainty associated with the same input, then this suggests epistemic uncertainty is a tool for OoD detection. Beyond a certain threshold of epistemic uncertainty, images which are out-of-distribution or highly unlikely will be more common. In this paper, we do not distinguish between OoD images and images which are \textit{extremely unlikely}, because in a practical setting the probability density function generating a complex set of images cannot be known precisely.

\subsection{Methods for OoD image generation}
\label{sec:ood_generation}

In the subsequent sections, OoD detection for epistemic uncertainty is investigated by running experiments involving datasets which are known to be out-of-distribution in reference to the amplitudes dataset. There are two different OoD image generation methods considered: (1) adding salt and pepper corruption to in-distribution data (an OoD noise source), and (2) using images from ImageNet (a different data generator).

\subsubsection{Salt and pepper corruption}
\label{sec:sp_corruption}

In salt and pepper (SP) corruption, random pixels in the image are filled with black (0) or white (1) pixels. Salt and pepper corruption is not a noise source present in the data generator, so sufficient corruption will yield out-of-distribution images. The top row of Figure \ref{fig:varying_ImNet_blend_and_SP_corruption} shows examples of varying levels of SP corruption. The progressive addition of salt and pepper noise enables experiments relying on variable levels of OoD noise in the image.


\subsubsection{Downsampled ImageNet images}
\label{sec:imagenet}

\begin{figure}[H]
	\centering
	\includegraphics[width=0.76\textwidth]{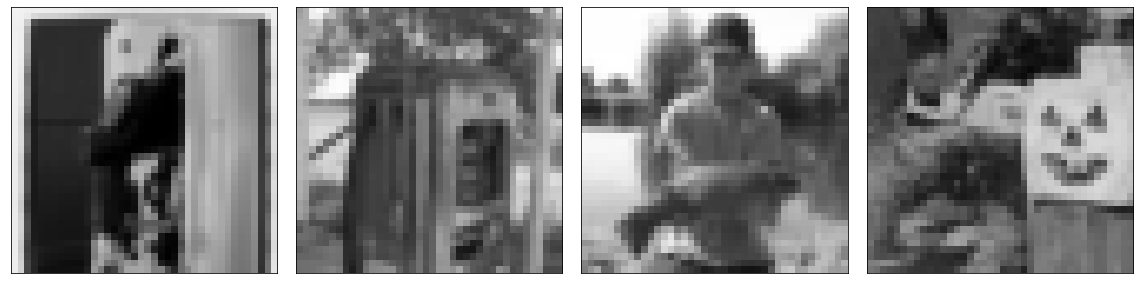}
	\caption{Examples of downsampled ImageNette images}
	\label{fig:ex_downsample_imagenette}
\end{figure}

ImageNet is a large database of images commonly used in computer vision tasks for benchmarking \citep{deng2009imagenet}. In our experiments with OoD detection, we use a smaller version of ImageNet called ImageNette created for the purpose of faster benchmarking and experiments \citep{imagewang}. The image generator for ImageNet is complex, very diverse, and has no closed-form generator, so ImageNet images are \textit{de facto} out-of-distribution relative to the amplitudes dataset. Figure \ref{fig:ex_downsample_imagenette} shows examples of the downsampled 40x40 ImageNet images.

\begin{figure}[H]
	\centering
	\includegraphics[width=0.80\textwidth]{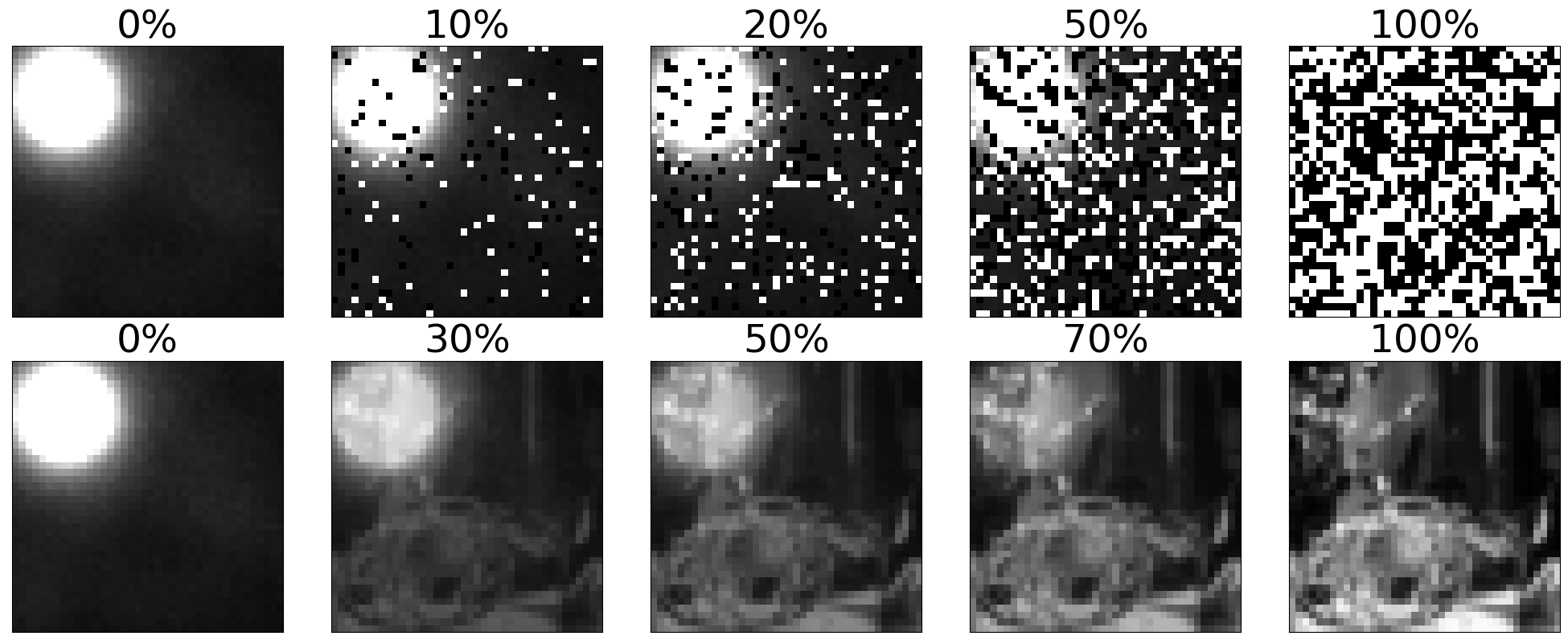}
	\caption{Varying levels of corruption to yield images which are progressively OoD. The top row shows a progressive corruption using the salt and pepper method, and the bottom row shows a progressive corruption using a gradual blending toward an ImageNet image.}
	\label{fig:varying_ImNet_blend_and_SP_corruption}
\end{figure}

Our method for creating images which are increasingly OoD using ImageNet was to use a linear blend of the iD data and the ImageNet data with $X_{new} = (1-\lambda)X_{iD} + \lambda X_{OoD}$ for $\lambda \in \{0, 0.1, 0.2, \dots 1.0\}$. In this blending, every in-distribution image gets paired with a random out-of-distribution image and that match stays consistent across all blend levels. The bottom row of Figure \ref{fig:varying_ImNet_blend_and_SP_corruption} shows an example slice of this blended dataset where an in-distribution image gets blended with an out-of-distribution image. The value of these progressively OoD datasets is the allowance for studying how UQ and OoD detection behaves when the datasets become more extreme in their distance from the iD image distribution.


\subsection{Epistemic uncertainty of out-of-distribution images}
\label{sec:diff_sigma_e}

This section provides empirical evidence that the typical distributions of $\hat{\sigma}_E$ differ on iD and OoD images. See Appendix \ref{app:pdf_inverse} for a more in-depth experiment linking high epistemic uncertainty in BNNs to a low value of the probability density function for an image.

\begin{figure}[H]
	\centering
	\includegraphics[width=1.0\textwidth]{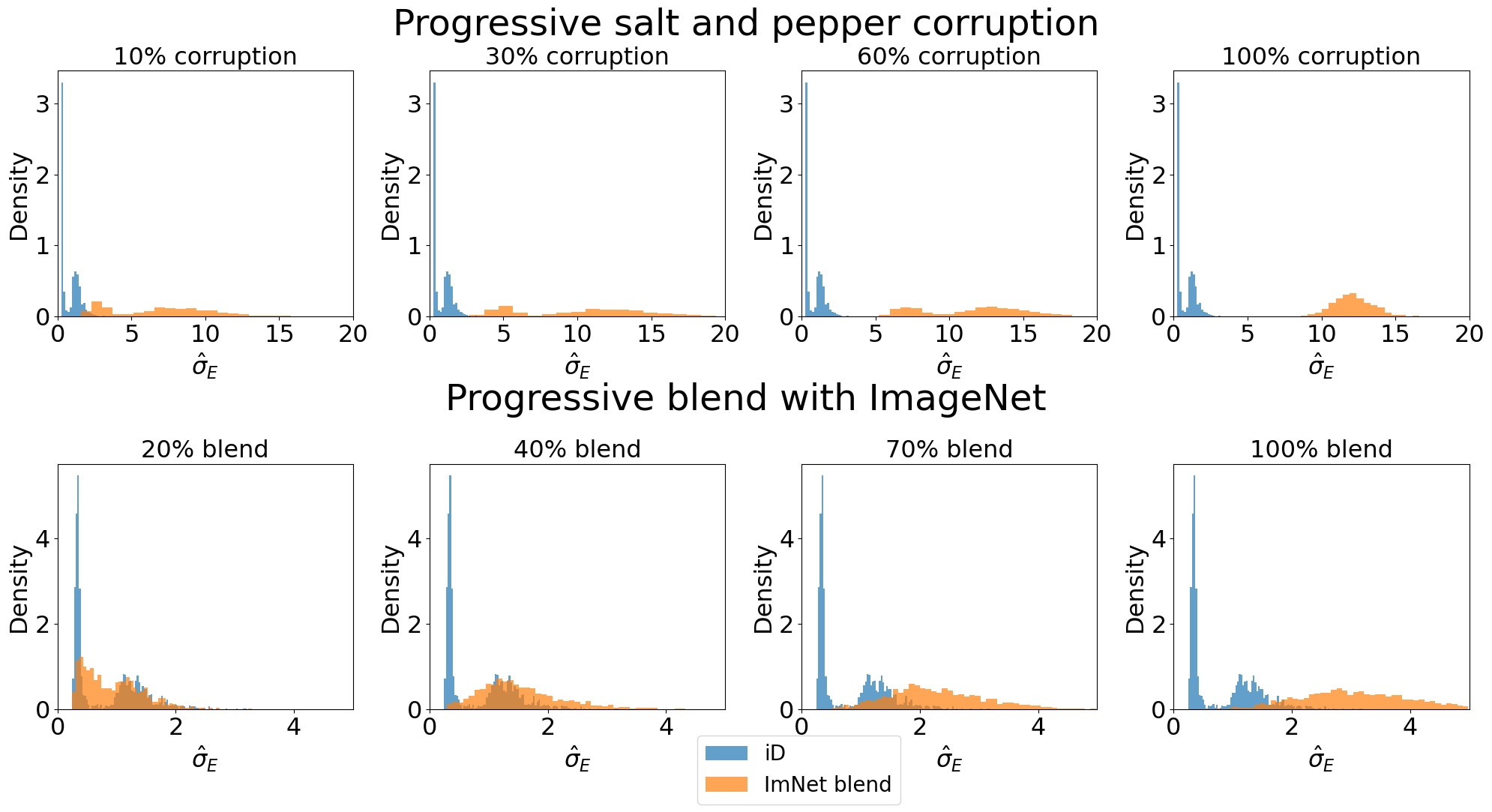}
	\caption{Differences in distribution of $\hat{\sigma}_E$ for a few different levels of corruption and blending. The top row shows the empirical distribution of $\hat{\sigma}_E$ for various levels of salt and pepper corruption, and the bottom row shows the empirical distribution of $\hat{\sigma}_E$ for various levels of blending with ImageNet.}
	\label{fig:separation_dist_SP_and_blend_levels}
\end{figure}

The separation between the distributions for $\hat{\sigma}_E$ on the amplitudes dataset for varying levels of salt and pepper corruption can be found in the top row of Figure \ref{fig:separation_dist_SP_and_blend_levels}. These results were generated by using 1500 in-distribution images from the test dataset and used architecture in Model 1 from Figure \ref{fig:MODEL_1} with a BNN resampling parameter of $T=100$ and training size of 10000 images. Notice that as the level of corruption increases, the differences in distribution of $\hat{\sigma}_E$ in the in-distribution data and the out-of-distribution increases as well. This suggests that as a set of images becomes increasingly out-of-distribution, the distinction between the distributions of $\hat{\sigma}_E$ becomes increasingly clear.


The bottom row of Figure \ref{fig:separation_dist_SP_and_blend_levels} shows the results for a similar experiment where 1500 in-distribution test images are linearly blended with a fixed out-of-distribution ImageNet image. This model used a BNN resampling parameter of $T=100$ and a training size of 10000 images. This figure also shows that as the influence of the OoD images increases, there is a more distinct separation in distribution of $\hat{\sigma}_E$. Notably, the separation in these figures is less stark than the separation in the top row of figures in Figure \ref{fig:separation_dist_SP_and_blend_levels}. The exact reason is unknown, but a possible explanation for the difference is that blended ImageNet images are soft and natural compared against the harshness of salt and pepper corruption. The softness may contribute to an increase in similarity to iD amplitude images, making it harder to detect an ImageNet image as being OoD than it is to detect a salt and pepper corrupted image as being OoD.


\subsection{An algorithm for OoD detection with epistemic uncertainty}
\label{sec:intrinsic_ood}

The differences in distributions in $\hat{\sigma}_E$ observed in Section \ref{sec:diff_sigma_e} suggest a simple algorithm for classifying a particular image as being in-distribution or out-of-distribution:

\begin{algorithm}
\caption{Algorithm for OoD detection with epistemic uncertainty} \label{alg:ood_epistemic}
\begin{algorithmic}
\Require Training set $\mathcal{D} = (\bm{X}, \bm{y})$, number of resamples $T \in \mathbb{N}$, false alarm rate $\alpha \in [0,1]$.
\State Train a BNN $\Phi: \mathcal{X} \to \mathcal{Y}$ with a training set $\mathcal{D} = (\bm{X}, \bm{y})$.
\State Resample the BNN $T$ times with an in-distribution validation set $\tilde{\mathcal{D}} = (\tilde{\bm{X}}, \tilde{\bm{y}})$ to obtain an estimate for $\hat{\sigma}_E$ for each validation image.
\State Set $RM$ to be the $100(1-\alpha)$ percentile of the empirical $\hat{\sigma}_E$ distribution from the validation set. Figure \ref{fig:ex_rejection_marks} shows example rejection marks for different levels of $\alpha$.
\State For a new image $X$, calculate $\hat{\sigma}_E$ and flag $X$ as out-of-distribution if $\hat{\sigma}_E > RM$ and flag $X$ as in-distribution otherwise.
\end{algorithmic}
\end{algorithm}

\begin{figure}[H]
	\centering
	\includegraphics[width=0.65\textwidth]{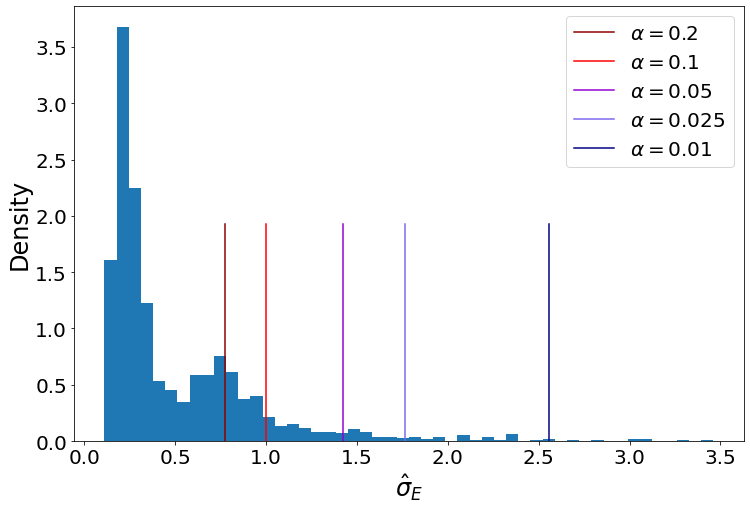}
	\caption{Example rejection marks for a few levels of $\alpha$. The model generating these predictions of $\hat{\sigma}_E$ for the test dataset is the same model from Section \ref{sec:diff_sigma_e}.}
	\label{fig:ex_rejection_marks}
\end{figure}

The $\alpha$ hyper parameter in Algorithm \ref{alg:ood_epistemic} can be thought of as a false alarm rate (equivalently, a type 1 error rate). If the validation set is an i.i.d. sample from the training distribution, then the algorithm is expected to classify $100 \times \alpha$\% of the in-distribution images as out-of-distribution. There can be no \textit{a priori} knowledge of the distribution of $\hat{\sigma}_E$ for out-of-distribution images encountered after model training, so \textit{any} method for out-of-distribution detection using epistemic uncertainty must rely on a similar threshold coming from in-distribution samples. 

\subsection{Factors influencing discrimination capability}
\label{sec:factors_in_separation}

Figure \ref{fig:separation_dist_SP_and_blend_levels} shows that as the images become increasingly OoD, the empirical separation in $\hat{\sigma}_E$ between iD and OoD images increases. There are other factors influencing the degree of separation in epistemic uncertainty, such as training set size, architecture of the BNN, number of training epochs, etc. This section investigates the effect of these variables on the level of empirical separation in $\hat{\sigma}_E$ between iD and OoD images using the salt and pepper corrupted dataset and blended ImageNet dataset as exemplar datasets. Sections \ref{sec:sep_training_size}, \ref{sec:sep_epochs}, and \ref{sec:sep_topology} show the results from experiments showing how separation changes as a function of these parameters where the results are averaged across five randomly initialized models to reduce run-to-run variance.

\subsubsection{Separation index (SI) - a measure for separation in distribution}

This section introduces the ``separation index (SI),'' a measure created to compare the difference between distributions of $\hat{\sigma}_E$ for in-distribution and out-of-distribution data. The SI is a useful way to quickly assess how a Bayesian neural network differentiates in-distribution and out-of-distribution data based upon $\hat{\sigma}_E$. Calculating the separation index is straightforward: the separation index for a set of OoD data is the proportion of the OoD data classified as OoD using the algorithm from Section \ref{sec:intrinsic_ood}. If the OoD detection method is working perfectly, then the SI will be 1.0 for the OoD data.

All experiments involving the SI will use a consistent value of $\alpha=0.05$. In the amplitudes dataset, there tended to be some in-distribution images registering abnormally high values of $\hat{\sigma}_E$, so it is safe to treat these instances as outliers. $\alpha=0.05$ was chosen to strike a balance between being robust to outliers in $\hat{\sigma}_E$ while maintaining a reasonable false alarm rate of 5\%. If desired, one could set $\alpha=0$ so that only images with $\hat{\sigma}_E$ higher than any iD values of $\hat{\sigma}_E$ would register as being OoD. In practice, the choice of $\alpha$ is made based on whether it is acceptable to reject outliers in the distribution as out-of-distribution with a given false alarm rate of $\alpha$. An optimal choice of $\alpha$ is problem-dependent and up to the user. This paper fixes $\alpha=0.05$ to simplify the comparison between empirical experiments.

\begin{figure}[H]
	\centering
	\includegraphics[width=0.75\textwidth]{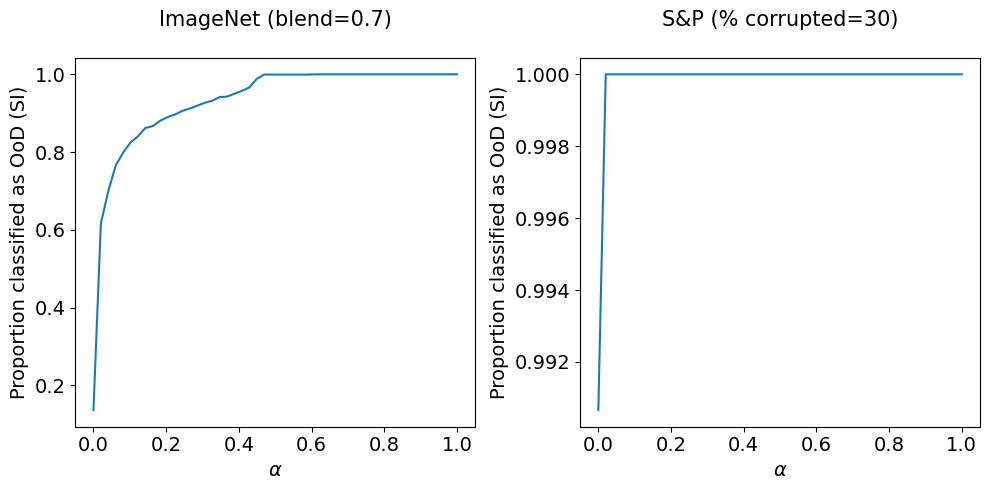}
	\caption{Instead of reporting SI for a fixed value of $\alpha$, ROC curves plotting the separation index (proportion of OoD data actually classified as OoD) as a function of $\alpha$ are an alternative way to assess the accuracy of a model.}
	\label{fig:alpha_ROC_curve_example}
\end{figure}

An alternative method to fixing a value of $\alpha$ is to plot the SI as a function of $\alpha$. This curve shows how the proportion of OoD data classified by the algorithm as OoD changes as a function of this parameter $\alpha$. These curves can be thought of as a receiver operating characteristic (ROC) curve. Figure \ref{fig:alpha_ROC_curve_example} shows an example curve for Model 1 discussed in Section \ref{sec:ex_alea_epis} on arbitrarily selected blended ImageNet and salt and pepper corrupted datasets. One potential metric for assessing the quality of the OoD discriminator is the area under this curve: an area of 0.50 would indicate a poor OoD detector making random decisions and an area of 1.0 would indicate a perfect OoD detector. However, in this paper we fix a value of $\alpha=0.05$ for the sake of simplicity and to avoid the clutter of reporting a large number of these plots.

\subsubsection{Separation as a function of training size}
\label{sec:sep_training_size}

\begin{figure}[H]
	\centering
	\includegraphics[width=0.95\textwidth]{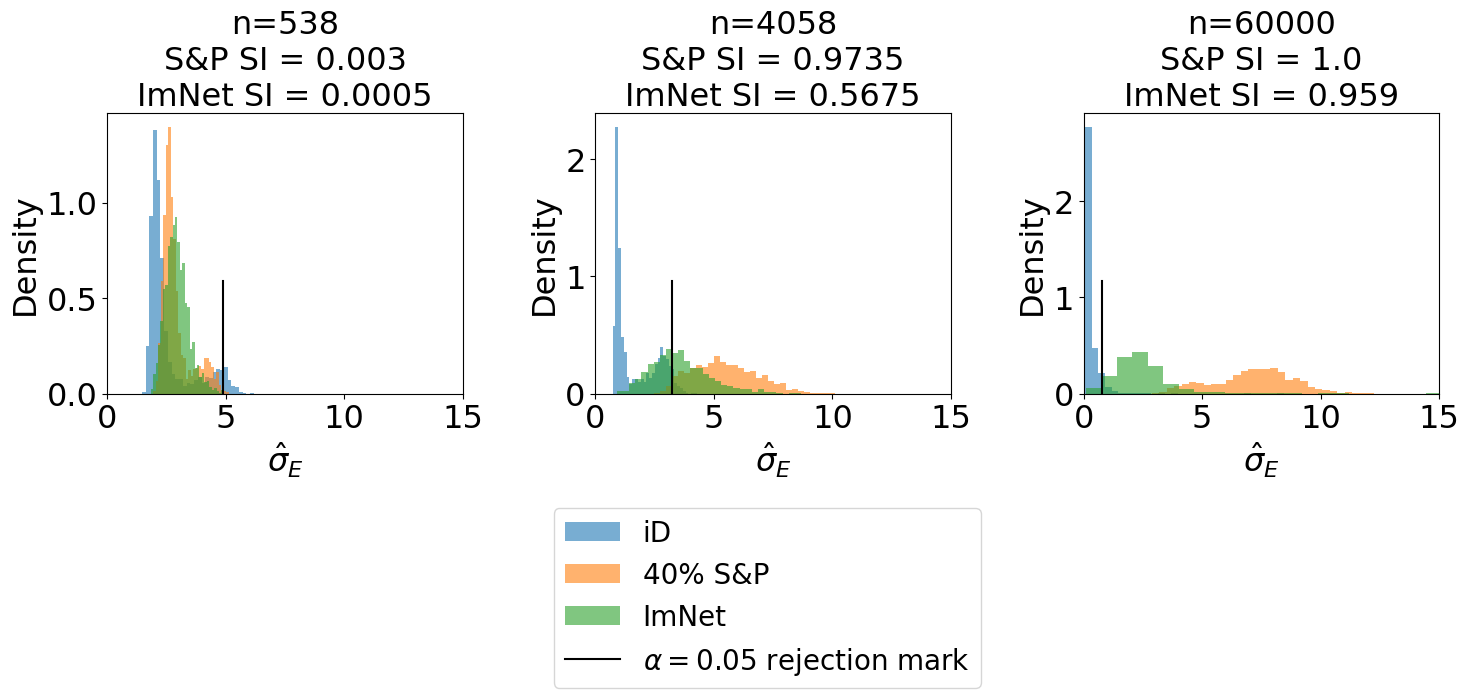}
	\caption{Examples of how separation in $\hat{\sigma}_E$ increases when sample size increases}
	\label{fig:SI_by_n_examples}
\end{figure}

As training size for a BNN increases, the separability in $\hat{\sigma}_E$ between iD and OoD data increases as well. Figure \ref{fig:SI_by_n_examples} empirically shows that as the training size $n$ increases, there is a bigger difference in the distribution of $\hat{\sigma}_E$ on the iD and two OoD datasets.

To evaluate the effect that training size has on the separation in predicted epistemic uncertainty between iD and OoD data for more than just three training set sizes, the separation index is calculated for varying levels of model training sizes. The in-distribution dataset has 500 images randomly selected from the test dataset, and the two out-of-distribution datasets both consist of 500 random images. The two out-of-distribution sets evaluated were the ImageNet dataset and a 40\% salt and pepper corrupted version of the iD test set. All models were trained with the architecture described in Figure \ref{fig:MODEL_1}. The training sample sizes were chosen to begin at 100, end at 60000, and be geometrically spaced with 20 total training set sizes.


\begin{figure}[H]
	\centering
	\begin{subfigure}[b]{0.45\textwidth}
		\centering
		\includegraphics[width=\textwidth]{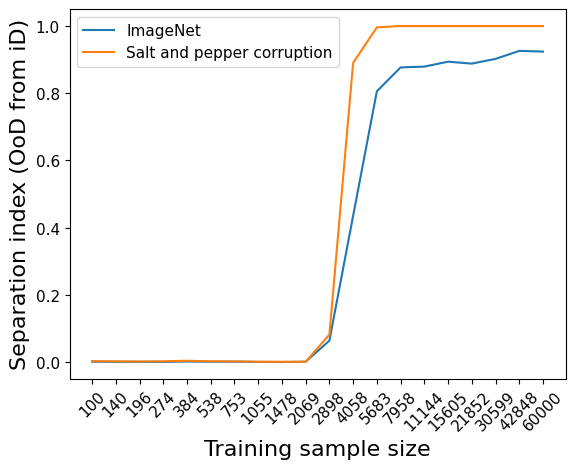}
		\caption{Separation index for iD and OoD data for varying training sizes.}
		\label{fig:avg_SI_by_training_size}
	\end{subfigure}
	\hfill
	\begin{subfigure}[b]{0.45\textwidth}
		\centering
		\includegraphics[width=\textwidth]{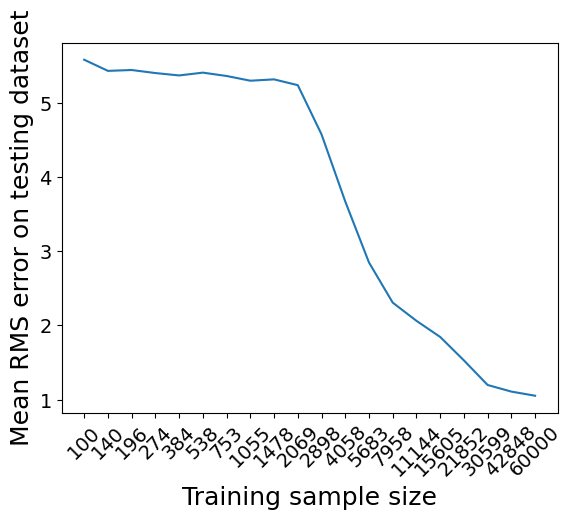}
		\caption{Average root mean squared error of the predicted amplitude against the true amplitude on the testing dataset.}
		\label{fig:test_RMS_by_training_size}
	\end{subfigure}
	\caption{Results as a function of a changing training sample size. In Figure \ref{fig:avg_SI_by_training_size}, we note an increase in separation index as the sample size increases, indicating that the capability for out-of-distribution detection increases as a function of sample size. Figure \ref{fig:test_RMS_by_training_size} shows that there is a corresponding decrease in average root mean squared error on the testing dataset as a function of sample size.}
	\label{fig:results_by_training_size}
\end{figure}

Figure \ref{fig:avg_SI_by_training_size} shows the separation index as a function of the training size. The plotted values are the average separation index for five different models (differing only in the random seed used for stochastic gradient descent) trained with the same sample size. This figure shows that as the training sample sizes increases, the separation in $\hat{\sigma}_E$ distribution for the iD and OoD data increases. This implies that OoD detection capability increases with training set size. Furthermore, Figure \ref{fig:test_RMS_by_training_size} shows a decrease in RMS error on the testing dataset with an the increase in sample size, so in this case there is not a trade-off between out-of-distribution detection capability and predictive accuracy.

\subsubsection{Separation as a function of training epochs}
\label{sec:sep_epochs}

\begin{figure}[H]
	\centering
	\includegraphics[width=0.95\textwidth]{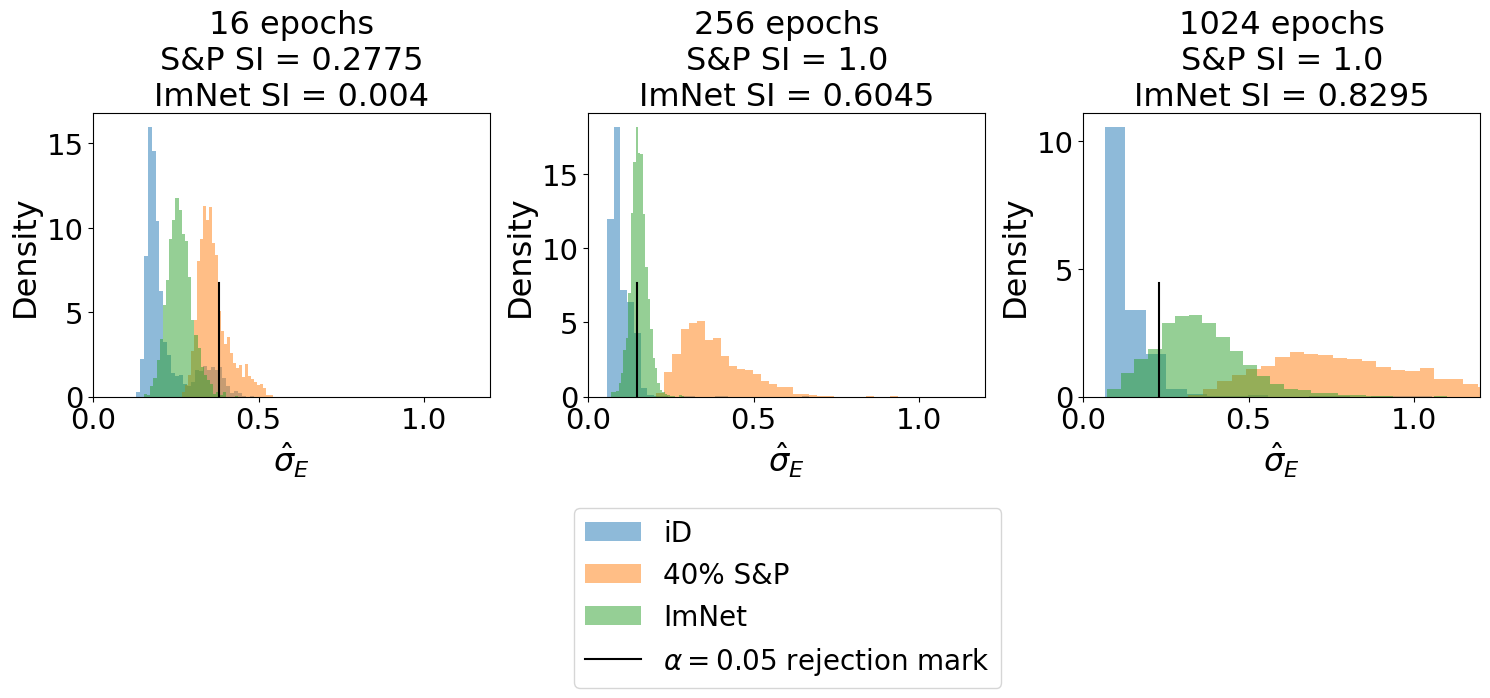}
	\caption{Examples separation in $\hat{\sigma}_E$ for a few different training epoch levels}
	\label{fig:SI_by_epochs_examples}
\end{figure}

As training epochs increases, so does the separation in distribution in $\hat{\sigma}_E$ on iD and OoD data. Figure \ref{fig:SI_by_epochs_examples} shows some examples of how the two class distributions begin to separate for a few different levels of training epochs. The in-distribution data is 500 images randomly selected from a test dataset, and there are two out-of-distribution datasets where 500 images are randomly selected. The first out-of-distribution set is the test dataset with 40\% of the pixels corrupted with salt and pepper corruption, and the second out-of-distribution set is 500 random ImageNet images.

\begin{figure}[H]
	\centering
	\begin{subfigure}[b]{0.45\textwidth}
		\centering
		\includegraphics[width=\textwidth]{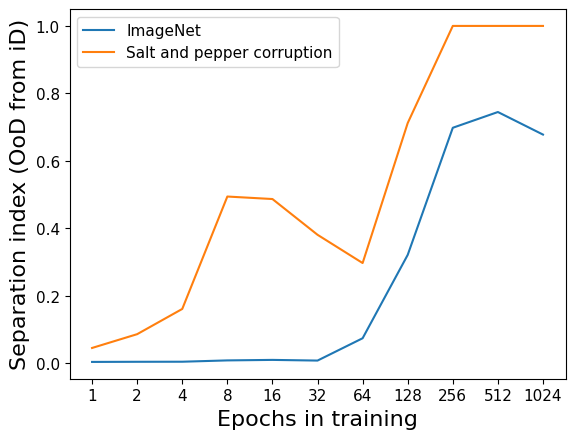}
		\caption{Separation index for iD and OoD data for a varying number of training epochs.}
		\label{fig:avg_SI_by_epochs}
	\end{subfigure}
	\hfill
	\begin{subfigure}[b]{0.45\textwidth}
		\centering
		\includegraphics[width=\textwidth]{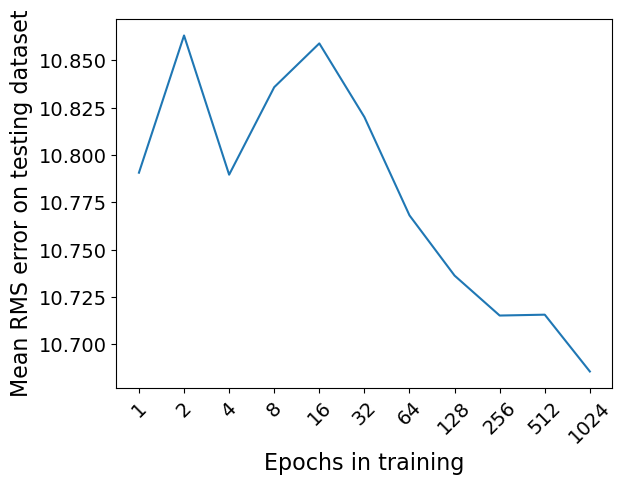}
		\caption{Average root mean squared error of the predicted amplitude against the true amplitude on the testing dataset.}
		\label{fig:test_RMS_by_epochs}
	\end{subfigure}
	\caption{Results as a function of a changing number of training size.}
	\label{fig:results_by_training_size}
\end{figure}

Figure \ref{fig:avg_SI_by_epochs} shows the separation index for the salt and pepper dataset and the ImageNet dataset against the in-distribution data as a function of the number of training epochs, using a model with architecture described in Figure \ref{fig:MODEL_1}. These results are for the average separation index over five randomly initialized models with the listed training set size.


Figure \ref{fig:avg_SI_by_epochs} shows that separation index increases with the number of training epochs. Furthermore, Figure \ref{fig:test_RMS_by_epochs} shows that average RMS error on the testing dataset decreases as training epochs increases, so in this setting there is not a tradeoff between the out-of-distribution detection capabilities of the BNN and its predictive accuracy on the testing dataset.

\subsubsection{Separation as a function of network topology}
\label{sec:sep_topology}

\begin{figure}[H]
	\centering
	\includegraphics[width=0.50\textwidth]{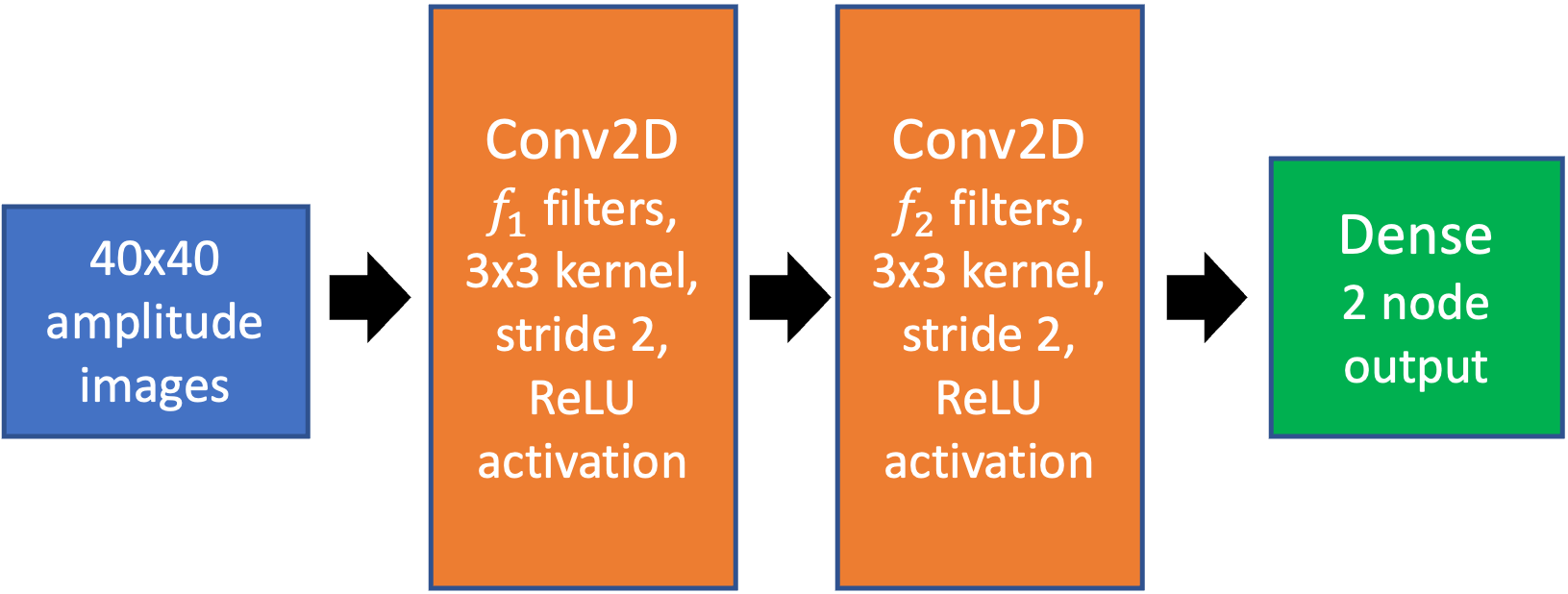}
	\caption{Architecture for Model 2. Note the similarity to Model 1, except that the number of filters in the 1st and 2nd convolutional layers is left variable. This model uses a batch size of 64, Adam optimizer learning rate of 0.0001, and 300 training epochs.}
	\label{fig:MODEL_4X}
\end{figure}

This section explores how changes in the topology of the convolutional neural network affect the separability in $\hat{\sigma}_E$. In particular, the number of convolutional filters in the first and second hidden layers is linked to separability. Figure \ref{fig:MODEL_4X} describes the architecture. The new model architecture is identical to the model from Figure \ref{fig:MODEL_1}, except that the number of filters in the first and second convolutional layers is left variable with sizes $f_1$ and $f_2$ respectively.

Figure \ref{fig:MODEL_4X} shows empirically that for different filter sizes, the differences in distribution of $\hat{\sigma}_E$ between iD and OoD data differ.

\begin{figure}[H]
	\centering
	\includegraphics[width=0.95\textwidth]{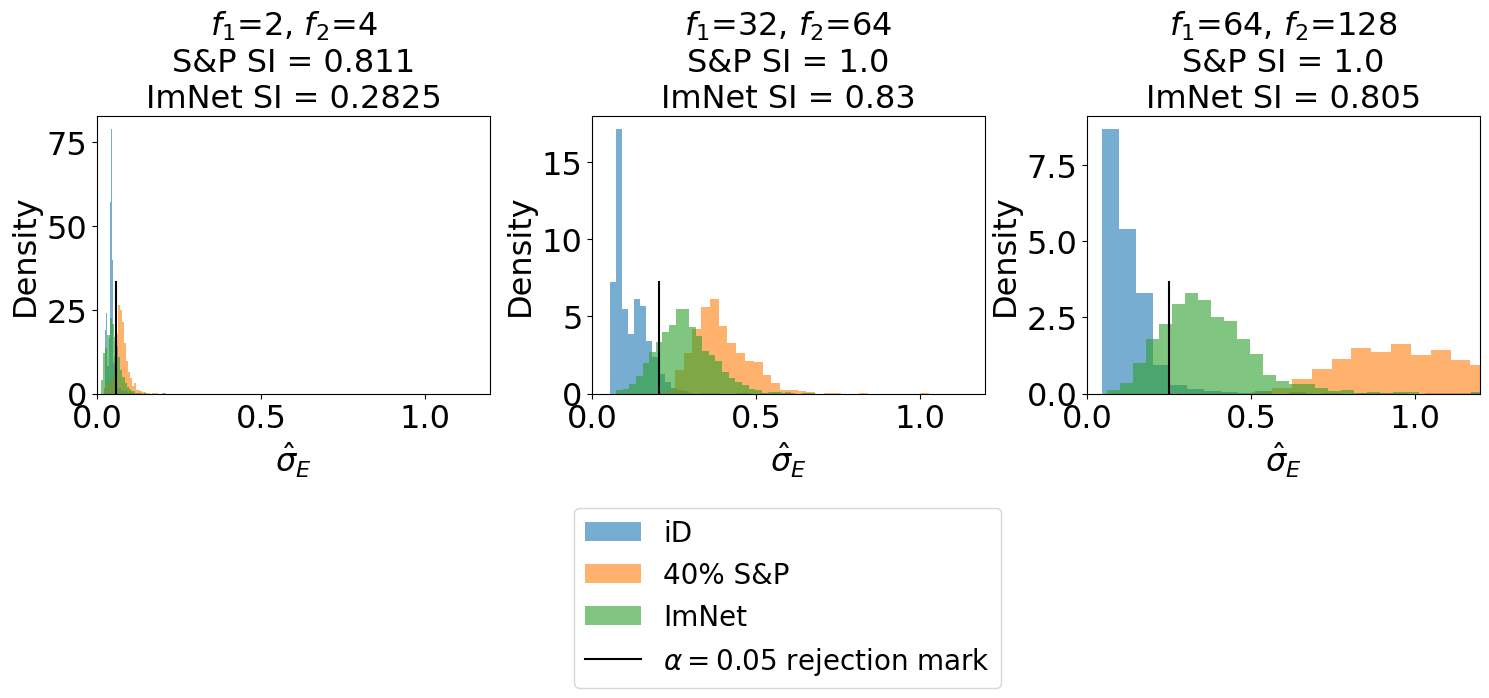}
	\caption{Examples of the different distributions of $\hat{\sigma}_E$ for iD and OoD data with a few different filter sizes $f_1$ and $f_2$.}
	\label{fig:SI_by_filtersize_examples}
\end{figure}

\begin{figure}[H]
	\centering
	\begin{subfigure}[b]{0.45\textwidth}
		\centering
		\includegraphics[width=\textwidth]{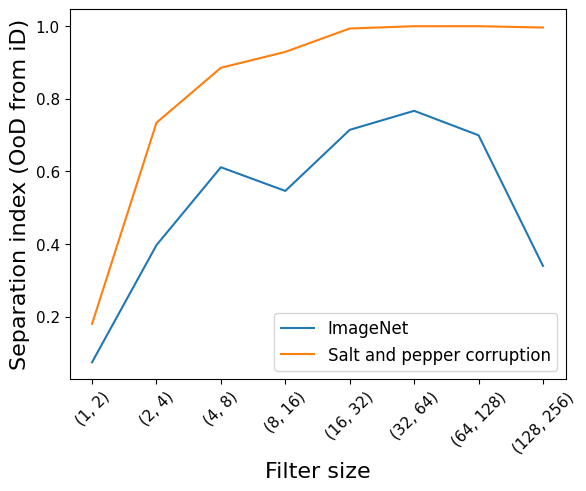}
		\caption{Separation index for OoD data and iD data for all filter sizes checked.}
		\label{fig:avg_SI_by_filtersizes}
	\end{subfigure}
	\hfill
	\begin{subfigure}[b]{0.45\textwidth}
		\centering
		\includegraphics[width=\textwidth]{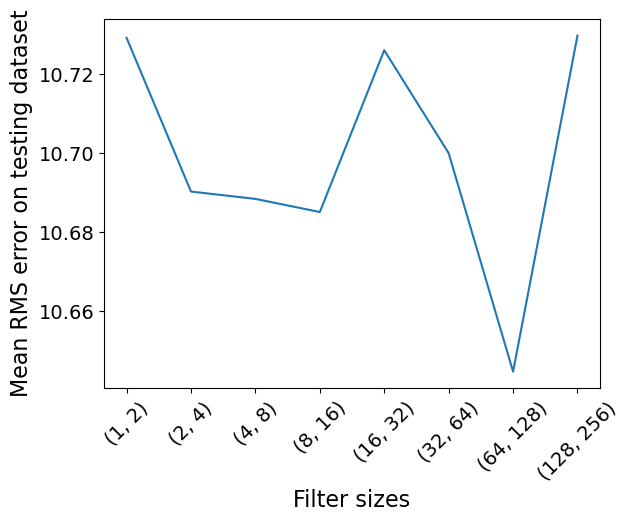}
		\caption{Average root mean squared error of the predicted amplitude against the true amplitude on the testing dataset for all filter sizes checked.}
		\label{fig:test_RMS_by_epochs}
	\end{subfigure}
	\caption{Results as a function of a changing network topology.}
	\label{fig:results_by_topology}
\end{figure}

Figure \ref{fig:avg_SI_by_filtersizes} shows the separation index for the two different OoD datasets for a few different levels of filter sizes. The results in this figure are the average separation index averaged over five different randomly initialized models with the listed filter sizes in the layers. This figure suggests that as the complexity of the network grows, class separability in $\hat{\sigma}_E$ tends to grow but there may be a sweet spot in model complexity where there are diminishing returns. This figure suggests that this sweet spot for this image set is in the range of $(f_1, f_2) = (8, 16)$ to $(32, 64)$.

Figure \ref{fig:test_RMS_by_epochs} shows noisy decrease in mean RMS error on the testing dataset as the filter size increases.

\subsection{Comparison of epistemic uncertainty algorithm with GAN discriminator}
\label{sec:comparison_with_GAN}

The OoD-detection feature described in Section \ref{sec:intrinsic_ood} is not the only neural network-based approach that one can take in OoD detection. Generative adversarial networks (GANs) can also be used for this task as well. A GAN is a system of two neural networks called the generator and discriminator \citep{goodfellow2014generative}. The generator is tasked with creating images that mimic the training images, and the discriminator is tasked with the binary classification problem of identifying if an image comes from the training set or comes from the generator. The loss functions for the generator and discriminator feed off each other so that the generator is rewarded when it successfully fools the discriminator, and the discriminator is rewarded when it is able to correctly identify a generated image as being fake. The result is a generator which is good at creating in-distribution images and a discriminator network good at detecting images which are out-of-distribution. This section compares the epistemic uncertainty approach described in Section \ref{sec:intrinsic_ood} with a trained discriminator network with a comparable model architecture.

\begin{figure}[H]
	\centering
	\includegraphics[width=0.80\textwidth]{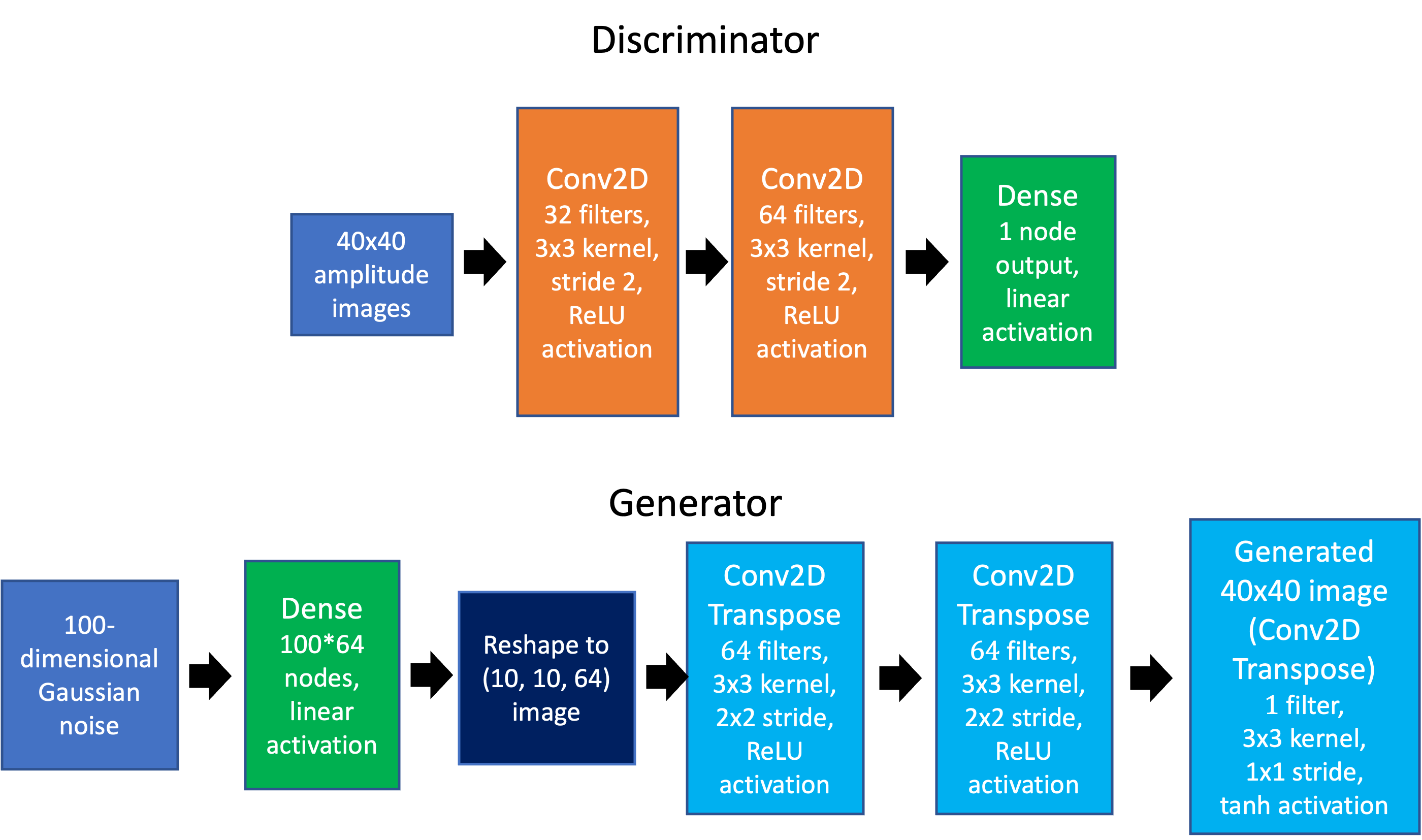}
	\caption{Architecture for the GAN. This model was trained with 600 training epochs, a learning rate of 0.0001, the Adam optimizer, and a batch size of 64.}
	\label{fig:MODEL_GAN32}
\end{figure}

Figure \ref{fig:MODEL_GAN32} shows the compared GAN model architecture. The GAN from this section is the only neural network model in this paper which is not a Bayesian neural network. The BNN architecture compared with the GAN is the model architecture in Figure \ref{fig:MODEL_1}. The GAN and the epistemic uncertainty BNN are trained on the same training dataset of 10000 images. All experiments in this section are averaged over five randomly initialized models to decrease the effect of run-to-run model variation in the results.

\subsubsection{Experiment 1: Detection rate by proportion of salt and pepper corruption}
\label{sec:sp_gan}

Recall the salt and pepper corrupted dataset introduced in Section \ref{sec:sp_corruption}. As the percentage of pixels corrupted increases, the images become further out-of-distribution. Therefore, it is expected that both the discriminator network and BNN approach to OoD detection will classify a greater portion of the data as being OoD if there is more salt and pepper noise. The experiments in this section link the proportion of image corruption with the proportion of the dataset that is classified as OoD by both the GAN and the epistemic uncertainty approach.

Figure \ref{fig:GAN_epis_SI_by_SP_corruption} shows the results for this experiment where the GAN discriminator and BNN have comparable architectures and learning parameters. The results for this figure are the average proportion classified as OoD over five different randomly initialized models with the same architecture. In this experiment, the GAN and the epistemic uncertainty approach for OoD detection yielded comparable results.


As expected, as the dataset becomes further out-of-distribution when salt and pepper corruption increases, both the GAN and the BNN classify an increasing proportion of the dataset as being OoD. In these results, the GAN and epistemic uncertainty approaches to OoD detection yield comparable OoD detection rates with the exception of a small window in the 5\% to 30\% salt and pepper corrupted images.

\subsubsection{Experiment 2: Blend with ImageNet}
\label{sec:imagenet_gan}

\begin{figure}
	\centering
	\begin{subfigure}[b]{0.45\textwidth}
		\centering
		\includegraphics[width=\textwidth]{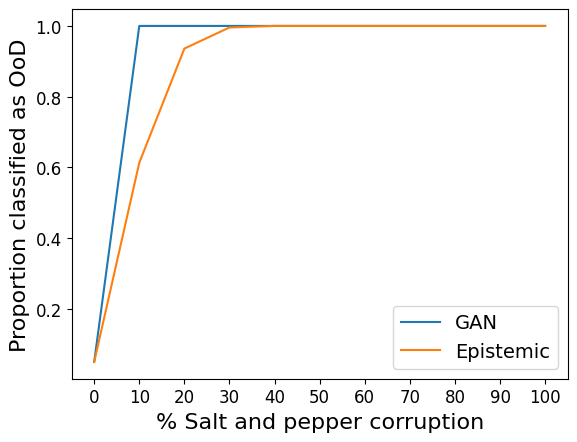}
		\caption{Proportion classified as OoD on the salt and pepper corrupted dataset.}
		\label{fig:GAN_epis_SI_by_SP_corruption}
	\end{subfigure}
	\hfill
	\begin{subfigure}[b]{0.45\textwidth}
		\centering
		\includegraphics[width=\textwidth]{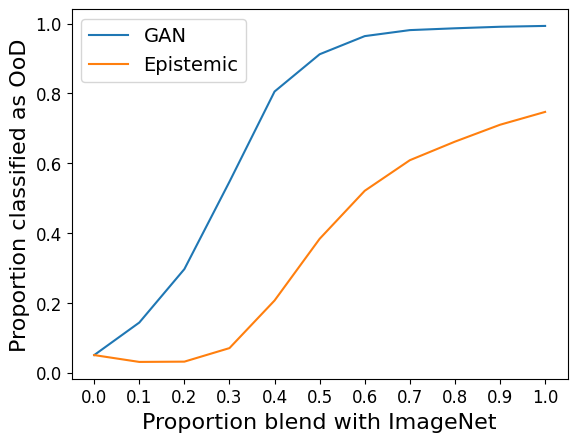}
		\caption{Proportion classified as OoD on the blended ImageNet dataset.}
		\label{fig:GAN_epis_SI_by_ImNet_blend}
	\end{subfigure}
	\caption{Proportion of the datasets classified as being OoD by both the GAN and the epistemic uncertainty approach as a function of the out-of-distribution level. As the images being increasingly OoD, both the GAN and the epistemic uncertainty approach classify more of the images as being OoD.}
	\label{fig:prop_ood_SP_BL}
\end{figure}

Recall the ImageNet dataset from Section \ref{sec:imagenet} where the in-distribution test dataset progressively blends with ImageNet images. Similar to the last section where the proportion of the set being classified as OoD increases with the amount of salt and pepper noise, the proportion of the dataset which gets classified as OoD increases as a function of the blend level. Figure \ref{fig:GAN_epis_SI_by_ImNet_blend} shows the result of the experiment where the $x$-axis shows the blending percentage for the ImageNet dataset, and the $y$-axis shows the percentage of the dataset which is classified as OoD by both the GAN and the epistemic uncertainty approach. The results are averaged over five different randomly initialized models with constant model architecture.


In this experiment, the epistemic uncertainty approach to OoD detection lags behind the GAN approach to OoD in terms of the proportion of OoD images which are classified as OoD. As seen in Figure \ref{fig:GAN_epis_SI_by_ImNet_blend}, the proportion of the ImageNet data classified as OoD as a function of the linear blend with ImageNet seems to plateau around 0.6 to 0.7. This indicates that roughly 60\% to 70\% of the ImageNet images are not classified as OoD by Algorithm \ref{alg:ood_epistemic}. A possible explanation could be that ImageNet images tend to be ``similar enough'' to amplitude images to where Algorithm \ref{alg:ood_epistemic} is simply not powerful enough to detect these differences in images.

Although the epistemic uncertainty approach to OoD detection is not as sensitive as the discriminator in a GAN, this experiment demonstrates that the epistemic uncertainty approach is \textit{useful}. Furthermore, the epistemic uncertainty approach has an advantage in that the OoD detection capability comes for free in a network which is also trained on the regression task of predicting the amplitude for the image, whereas the discriminator in the GAN has no ability to predict amplitude and requires a complementary secondary network.

\section{Conclusion and future work}

In summary, for the example dataset introduced in Section \ref{sec:data_description} we showed that epistemic uncertainty differs between in-distribution and out-of-distribution images for a BNN trained on only in-distribution images. This difference in distributions directly inspired Algorithm \ref{alg:ood_epistemic}, a simple algorithm for OoD detection using this difference in distribution of epistemic uncertainty. From Section \ref{sec:factors_in_separation}, we provided empirical studies that demonstrate the separation in distribution of epistemic uncertainty as a function of training size, the number of training epochs, and the model architecture. Concluding this paper with a novel comparison of Algorithm \ref{alg:ood_epistemic} with GANs, we showed that the OoD detection capability of Algorithm \ref{alg:ood_epistemic} lags slightly behind that of the discriminator network in a GAN, but still produces results that indicate Algorithm \ref{alg:ood_epistemic} is useful for OoD detection.

Future work may involve investigating how the choice of prior distribution over weights in the BNN affects OoD detection capability, as all networks trained in this paper were trained using a prior distribution of independent standard normal distributions. Other potential avenues of work include investigating adversarial attempts against OoD detection with BNNs, in the same way that others have investigated adversarial attempts against discriminator networks in GANs \citep{zhou2018don}.

\section{Acknowledgements}

We would like to thank Teresa Portone at Sandia National Laboratories for providing a technical review and helpful suggestions for this paper. 

Funds for this work were provided by the US Department of Energy, Workforce Development for Teachers and Scientists, as well as the Sandia LDRD office.

Sandia National Laboratories is a multi-mission laboratory managed and operated by National Technology and Engineering Solutions of Sandia, LLC., a wholly owned subsidiary of Honeywell International, Inc., for the U.S. Department of Energy’s National Nuclear Security Administration under contract DE-NA-0003525.

This paper describes objective technical results and analysis. Any subjective views or opinions that might be expressed in the paper do not necessarily represent the views of the U.S. Department of Energy or the United States Government.

\bibliography{ref} 

\newpage
\appendix

\section{Epistemic uncertainty is inversely proportional to density function}
\label{app:pdf_inverse}

The reference \citep{oaza1996bayesian} states that the epistemic uncertainty in Bayesian methods at an input is approximately inversely proportional to the value of the probability density function at that input. If true, this justifies the epistemic uncertainty driven approach to out-of-distribution detection introduced in Section \ref{sec:intrinsic_ood} because out-of-distribution images will have low probability density function values, hence high values of epistemic uncertainty. Section \ref{sec:diff_sigma_e} provides empirical evidence that for some types of out-of-distribution images, the value of epistemic uncertainty in a trained BNN is higher than for in-distribution images. Note that in a practical setting, the probability density function is impossible to know exactly, so for the purposes of this paper there is no distinction between out-of-distribution images and extremely unlikely (i.e., very low values of probability density function) images.

This section shows a simple example for a $2 \times 2$ pixel image generator where the distributions for the pixels are known. This allows for an evaluation of the claim that the BNN epistemic uncertainty for an image is approximately inversely proportional to the probability density function (PDF) evaluated at that image, because the exact PDF of the images are known.

\begin{figure}[H]
	\centering
	\begin{subfigure}[b]{0.50\textwidth}
		\centering
		\includegraphics[width=\textwidth]{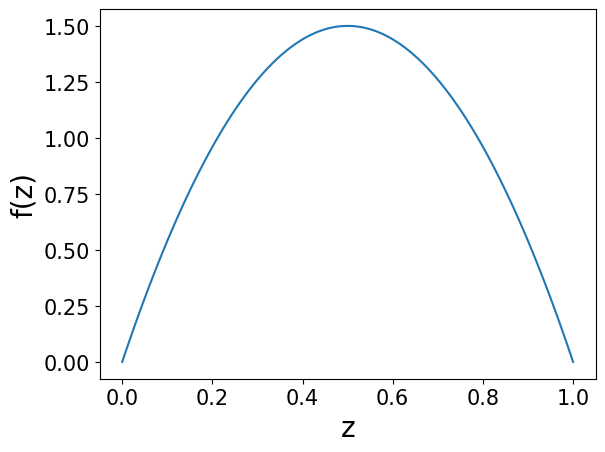}
		\caption{Probability density function for the beta distribution generating each pixel in the images}
		\label{fig:pdf_beta}
	\end{subfigure}
	\hfill
	\begin{subfigure}[b]{0.35\textwidth}
		\centering
		\includegraphics[width=\textwidth]{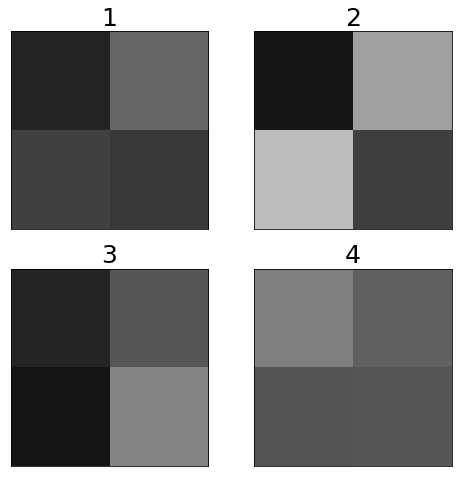}
		\caption{Example images from the data generation process}
		\label{fig:ex_beta_im}
	\end{subfigure}
	\caption{Distribution of pixels in image and example images}
	\label{fig:beta_generation_process}
\end{figure}

In this experiment, the pixels in the $2 \times 2$ are independent and identically distributed according to a $\text{Beta}(\alpha=2, \beta=2)$ distribution, the distribution shown in Figure \ref{fig:pdf_beta}. Because the generation of each pixel is independent, it is straightforward to find the value of the probability density function at each image by multiplying the probability density function values at each pixel together. Example images from this data generation process are shown in Figure \ref{fig:ex_beta_im}.

\begin{figure}[H]
	\centering
	\includegraphics[width=0.30\textwidth]{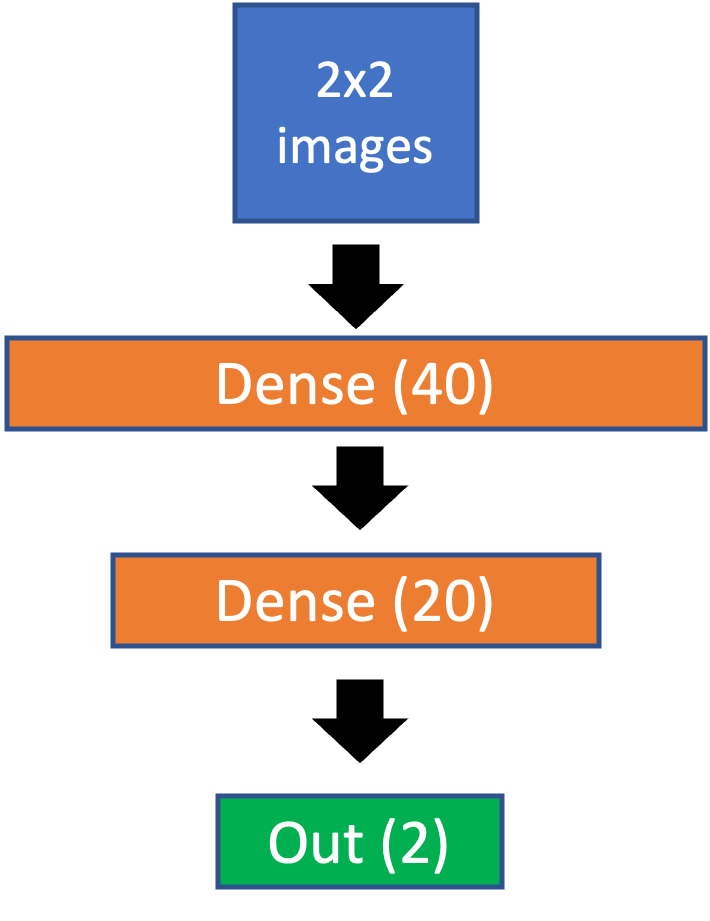}
	\caption{Model architecture for the $2 \times 2$ pixel image experiment}
	\label{fig:MODEL_pdf_inverse_prop}
\end{figure}

\begin{figure}[H]
	\centering
	\includegraphics[width=0.60\textwidth]{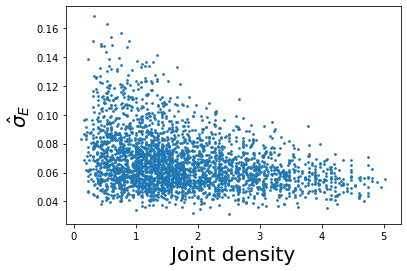}
	\caption{Scatter plot of $\hat{\sigma}_E$ from BNN and probability density function for images in the sampled mesh grid dataset}
	\label{fig:mesh_grid_density_vs_sigmaE}
\end{figure}

\begin{figure}[H]
	\centering
	\includegraphics[width=0.66\textwidth]{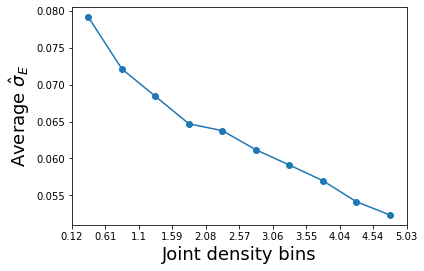}
	\caption{Average value of $\hat{\sigma}_E$ from equally spaced bins of joint density values in Figure \ref{fig:mesh_grid_density_vs_sigmaE}.}
	\label{fig:AVG_mesh_grid_density_vs_sigmaE}
\end{figure}

The epistemic uncertainty driven approach to OoD detection relies on a supervised learning task, so this simulation uses an arbitrarily-chosen linear labeling function:
\[Y = -2X_{1,1} + 3X_{1,2} -4X_{2,1} + 8X_{2,2} + \epsilon\]
\[\epsilon \sim \mathcal{N}(0, 0.5^2)\]
With this labeling function, we trained a BNN with 5000 training images, using the model architecture in Figure \ref{fig:MODEL_pdf_inverse_prop}. This network used a Gaussian negative-log likelihood loss function, predicting both $\hat{y}$ and $\hat{\sigma}_A$ as described in Section \ref{sec:calc_alea}. The network was trained with a batch size of 64, a learning rate of 0.01, and 150 epochs.
The probability density function at a variety of points is evaluated by creating an array of 20 equally spaced points between 0 and 1, and expanding this along 4 dimensions to create $20^4=160000$ equally spaced points in the space $[0,1] \times [0,1] \times [0,1] \times [0,1]$. Then, 3000 images are randomly sampled from the 160000 points to use as the test dataset. The evaluated probability density function and BNN epistemic uncertainty is calculated for each of these images. A scatter plot for all these results is contained in Figure \ref{fig:mesh_grid_density_vs_sigmaE}.
For a clearer depiction of the average trend, Figure \ref{fig:AVG_mesh_grid_density_vs_sigmaE} shows the average value of $\hat{\sigma}_E$ for equally spaced bins of joint density values.
This experiment shows that for a BNN, there is evidence that the predicted epistemic uncertainty for an input is approximately inversely proportional to the probability density function evaluated at the image. In a practical setting the true probability density function value for an image from a complex data generator (e.g., ImageNet) is unknown, but rare events (or images which are out-of-distribution) should have a low probability density function value accompanied by a higher value of predicted epistemic uncertainty.

\section{Evidence of heteroscedastic noise in amplitudes dataset}
\label{app:heteroscedastic_evidence}

In Section \ref{sec:calc_alea}, predicted aleatoric uncertainty $\hat{\sigma}_A$ comes from the output of a neural network, so it is dependent on the image $X$. Allowing $\hat{\sigma}_A$ to depend on the image implicitly admits a heteroscedastic noise model, a model where the extremity of the noise is assumed to be non-constant and dependent on the particular input to the model. This section argues that a heteroscedastic noise model is a good assumption for the amplitudes dataset introduced in this paper.

Although the noise distribution for the pixels in the image generator is known, the exact noise distribution for amplitude as a function of an image is unknown because the noise distribution for the pixels is applied to the image after presetting an amplitude for the image. Although the noise distribution for amplitude as a function of an image is unknown, a heteroscedastic noise model is still appropriate due to certain empirical observations in the data. In the amplitudes dataset, some wide ranges of amplitudes have little difference in images (e.g., amplitude ranges which fully saturate the image with the event), and in other narrower ranges small perturbations in amplitude correspond to much different images. An image which is fully saturated inherently contains more uncertainty in the exact amplitude due to the wide range of amplitude values which can fully saturate an image. For an image with a smaller event where the event fully lies within the image, there is a much smaller range of amplitudes which could lead to that particular image, so there is less uncertainty inherent in the data.

Another argument for a heteroscedastic noise model is the observation that heteroscedastic noise models lead to well-calibrated confidence intervals. A well-calibrated model in the context of confidence intervals is a model where a $100(1-\alpha)$\% confidence interval truly contains $100(1-\alpha)$\% of the data. This can be evaluated by creating confidence intervals on a test dataset for a variety of confidence levels, and then checking that the confidence level matches well with the percentage of the true test dataset labels which lie in their respective confidence intervals. To construct a $100(1-\alpha)$\% confidence interval with a BNN using the techniques described in this paper, the calculation $\hat{y} \pm z_{\alpha/2} \cdot \hat{\sigma}_A$ is used where $z_{\alpha/2}$ is the inverse standard normal Gaussian CDF evaluated at $\alpha/2$. This particular confidence interval calculation comes from the Gaussian noise model assumption for the true label $Y$.

\begin{figure}[H]
	\centering
	\includegraphics[width=0.70\textwidth]{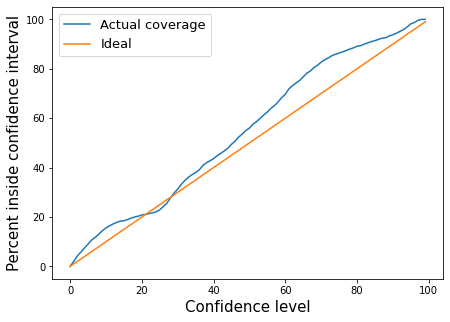}
	\caption{An example coverage plot demonstrating well-calibrated confidence intervals.}
	\label{fig:appendixB_coverage}
\end{figure}

This confidence interval evaluation lends itself to a visual check called a ``calibration plot,'' a type of plot where confidence levels are plotted on the $x$-axis and the percentage of true labels in the confidence intervals are on the $y$-axis. For a model with perfect calibration, the coverage plot will closely follow a $y=x$ line on the plot. Figure \ref{fig:appendixB_coverage} shows an example coverage plot for a model which is trained to predict aleatoric sigma with a heteroscedastic noise model.

\begin{figure}[H]
	\centering
	\includegraphics[width=0.55\textwidth]{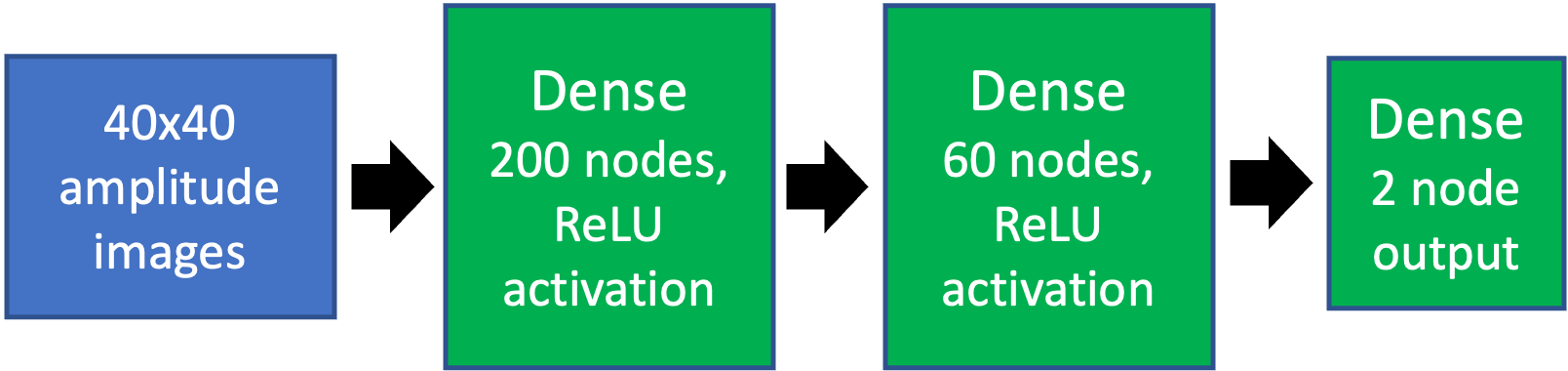}
	\caption{Model architecture used for the coverage plot above. Training used the Adam optimizer with a learning rate of 0.0005, 600 epochs, and a batch size of 64.}
	\label{fig:MODEL_appB}
\end{figure}

The model architecture used to generate the calibration curve in Figure \ref{fig:appendixB_coverage} can be found in Figure \ref{fig:MODEL_appB}. One will notice that the model architecture in this diagram differs from the convolutional neural network architecture used throughout this paper. The reasoning for this is that it was empirically more difficult to get well-calibrated confidence intervals from $(\hat{y}, \hat{\sigma}_A)$ when using the convolutional neural network architecture. The intent of the model used in this appendix was purely to show that a heteroscedastic noise model for $\hat{\sigma}_A$ was appropriate, and densely connected neural networks seem to be better for this task. 

However, convolutional neural network architectures were significantly better than dense neural networks in creating a separation in $\hat{\sigma}_E$ distribution between in-distribution and out-of-distribution data. A heavy emphasis on this paper is placed on out-of-distribution detection using this separation in $\hat{\sigma}_E$, and well-calibrated confidence intervals are not needed to do this task well. This appendix is included for completion to show that the approach to calculating $\hat{\sigma}_A$ from Section \ref{sec:calc_alea} is justified.

Note also that the learning rate in this network differs from the learning rate of 0.0001 used throughout the rest of this paper. This learning rate was chosen on an empirical basis so that the dense neural network architecture described above worked well with the task.

\section{Aleatoric and epistemic uncertainty as a function of image noise}
\label{app:iD_noise_levels}

This section provides an extra empirical study lending credibility to the notions and definitions of aleatoric and epistemic uncertainty introduced in Section \ref{sec:calculating_uncertainty}. If aleatoric uncertainty really does capture the uncertainty inherent in the data itself, then increasing the level of background noise in the image generator itself will increase the variability inherent in the problem and make it harder to detect a signal. Similarly, if epistemic uncertainty captures the uncertainty in the model parameters then increasing the level of background noise in the image generator should not have much effect on this level of uncertainty, because the levels of noise seen in the test set are also seen in the training set, so there is no extra uncertainty being imbued into the model parameters by changing the nature of the image generator itself.

\begin{figure}[H]
	\centering
	\begin{subfigure}[b]{0.45\textwidth}
		\centering
		\includegraphics[width=\textwidth]{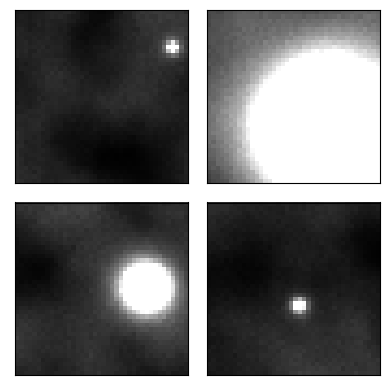}
		\caption{Four example images from the dataset generator with the background noise level set to 512.}
		\label{fig:example_noise512_images}
	\end{subfigure}
	\hfill
	\begin{subfigure}[b]{0.45\textwidth}
		\centering
		\includegraphics[width=\textwidth]{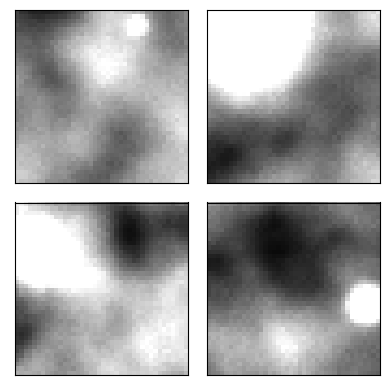}
		\caption{Four example images from the dataset generator with the background noise level set to 2048.}
		\label{fig:example_noise2048_images}
	\end{subfigure}
		
	\caption{Example images from the dataset generator with different levels of background noise.}
	\label{fig:example_background_noise}
\end{figure}

One toggle available in the amplitudes image generator introduced in Section \ref{sec:data_description} is the level of background noise in the image. The generator uniformly picks a noise scaling factor and then applies this scaling factor to a noise pattern in the image before superimposing the event image on top. As a demonstration of this toggle, Figure \ref{fig:example_noise512_images} shows example images coming from a background noise level set to 512, and Figure \ref{fig:example_noise2048_images} shows example images coming from a background noise level of 2048.

\begin{figure}[H]
	\centering
	\includegraphics[width=0.85\textwidth]{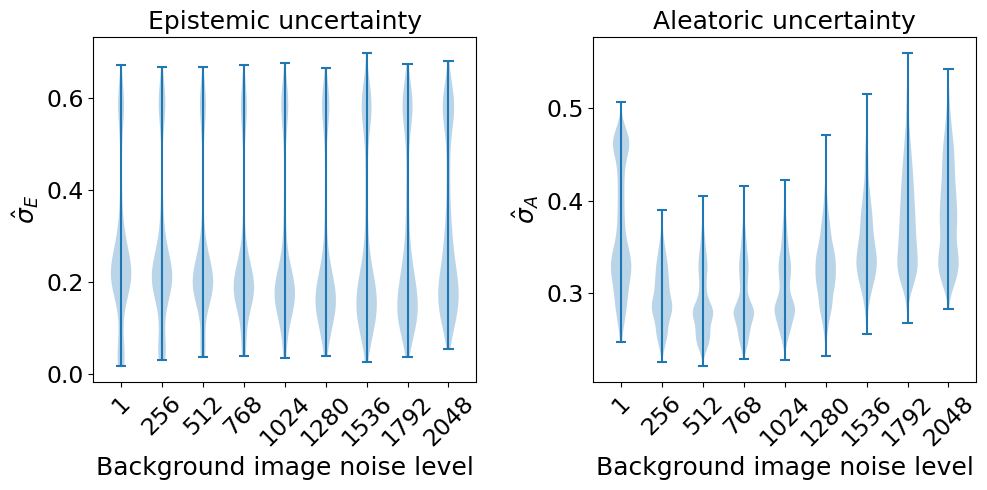}
	\caption{Distributions of aleatoric and epistemic uncertainty for different in-distribution noise levels.}
	\label{fig:violin_ae_iD_noise}
\end{figure}

A model was trained on an image dataset of 15000 images containing background noise levels uniformly picked from 1 to 2048 so that all noise levels were represented in the training dataset. The distribution of aleatoric and epistemic uncertainty was recorded for testing datasets where the noise levels are all set to exactly 1, 256, 512, 768, 1024, 1280, 1536, 1792, and 2048. The resulting distributions of aleatoric and epistemic uncertainty are recorded in Figure \ref{fig:violin_ae_iD_noise}.

Figure \ref{fig:violin_ae_iD_noise} shows that as a general rule, epistemic uncertainty stays constant as the level of background noise increases, and aleatoric uncertainty increases as the level of in-distribution noise increases. One strange feature of the aleatoric uncertainty in Figure \ref{fig:violin_ae_iD_noise} is the abnormally high level of aleatoric uncertainty in the images with a background noise level of 1. One potential explanation is that by cutting out all background noise, there is significant homogeneity in this image dataset, so if by chance one of the images has a high level of $\hat{\sigma}_A$, then many of the others will also concentrate in that spot. Regardless of the strange feature noted at a background noise level of 1, the trend of increasing $\hat{\sigma}_A$ as a function of background image noise level is noteworthy.

This experiment shows that our notions of aleatoric and epistemic uncertainty really do capture the respective notions of uncertainty inherent in the dataset and uncertainty in the model parameters.

\end{document}